\documentclass[sigconf]{aamas}

\usepackage{balance} 

\usepackage{hyperref}
\usepackage{url}
\usepackage{booktabs}       
\usepackage{amsfonts}       
\usepackage{nicefrac}       
\usepackage{microtype}      
\usepackage{xcolor}         
\usepackage{amsmath, amsthm}
\usepackage{graphicx}
\usepackage{subfigure}
\usepackage{enumitem}
\usepackage{wrapfig}
\usepackage{algorithmicx, algorithm, algpseudocode}

\newtheorem{theorem}{Theorem.}

\renewcommand\footnotetextcopyrightpermission[1]{} 
\acmConference[]{}{}{} 
\acmBooktitle{}
\acmPrice{}
\acmISBN{}
\acmDOI{}

\title[Formatting Instructions]{Mitigating Distribution Shift in Model-based Offline RL via Shifts-aware Reward Learning}

\author{Wang Luo}
\affiliation{
  \institution{UCAS}
  \country{China}}
\email{luowang21@mails.ucas.ac.cn}

\author{Haoran Li}
\affiliation{
  \institution{UCAS}
  \country{China}}

\author{Zicheng Zhang}
\affiliation{
  \institution{UCAS}
  \country{China}}

\author{Congying Han}
\affiliation{
  \institution{UCAS}
  \country{China}}

\author{Chi Zhou}
\affiliation{
  \institution{UCAS}
  \country{China}}

\author{Jiayu Lv}
\affiliation{
  \institution{UCAS}
  \country{China}}

\author{Tiande Guo}
\affiliation{
  \institution{UCAS}
  \country{China}}

\begin{abstract}
Model-based offline reinforcement learning trains policies using pre-collected datasets and learned environment models, eliminating the need for direct real-world environment interaction. 
However, this paradigm is inherently challenged by distribution shift~(DS). 
Existing methods address this issue by leveraging off-policy mechanisms and estimating model uncertainty, 
but they often result in inconsistent objectives and lack a unified theoretical foundation. 
This paper offers a comprehensive analysis that disentangles the problem into two fundamental components: model bias and policy shift. 
Our theoretical and empirical investigations reveal how these factors distort value estimation and restrict policy optimization.
To tackle these challenges, we derive a novel shifts-aware reward through a unified probabilistic inference framework, which modifies the vanilla reward to refine value learning and facilitate policy training. 
Building on this, we develop a practical implementation that leverages classifier-based techniques to approximate the adjusted reward for effective policy optimization.
Empirical results across multiple benchmarks demonstrate that the proposed approach mitigates distribution shift and achieves superior or comparable performance, validating our theoretical insights.
\end{abstract}


\newcommand{\BibTeX}{\rm B\kern-.05em{\sc i\kern-.025em b}\kern-.08em\TeX}

\begin{document}


\pagestyle{fancy}
\fancyhead{}


\maketitle 


\section{Introduction}
Offline reinforcement learning~(RL)~\cite{lange2012batch,levine2020offline} learns policies from the offline dataset generated by a behavior policy, avoiding additional online interaction with the environment. This approach shows remarkable potential in data-driven decision scenarios \cite{emerson2023offline, sinha2022s4rl} where exploration is costly or hazardous.
The model-based framework~\cite{luo2018algorithmic} is particularly effective for offline RL~\cite{yu2020mopo,yu2021combo,sun2023model,li2024settling}. It involves learning models of the environment from the offline dataset and utilizing these models to generate data for policy training. 
However, directly learning a policy using offline and synthetic data introduces the distribution shift challenge. 
This shift causes the training objective to significantly deviate from the true objective of reinforcement learning, leading to poor performance during testing.


Prior methods in model-based offline RL focus on directly leveraging off-policy methods in online RL and heuristic model uncertainty for conservative learning to mitigate the impact of distribution shift. 
For instance, MOPO~\cite{yu2020mopo} uses the predicted variance of the learned model as an uncertainty penalty on rewards.
MOBILE~\cite{sun2023model} penalizes value learning of model data according to the uncertainty of the Bellman Q-function estimated by ensemble models. 
RAMBO~\cite{rigter2022rambo} considers the trajectory distribution in the worst case and adversarially trains the policy and model while ensuring accurate predictions.
However, these approaches result in a biased objective by only addressing the distribution shift from model inaccuracies, leaving the broader distribution shift unresolved.

\begin{figure*}[t]
\centering
\includegraphics[width=0.9\linewidth]{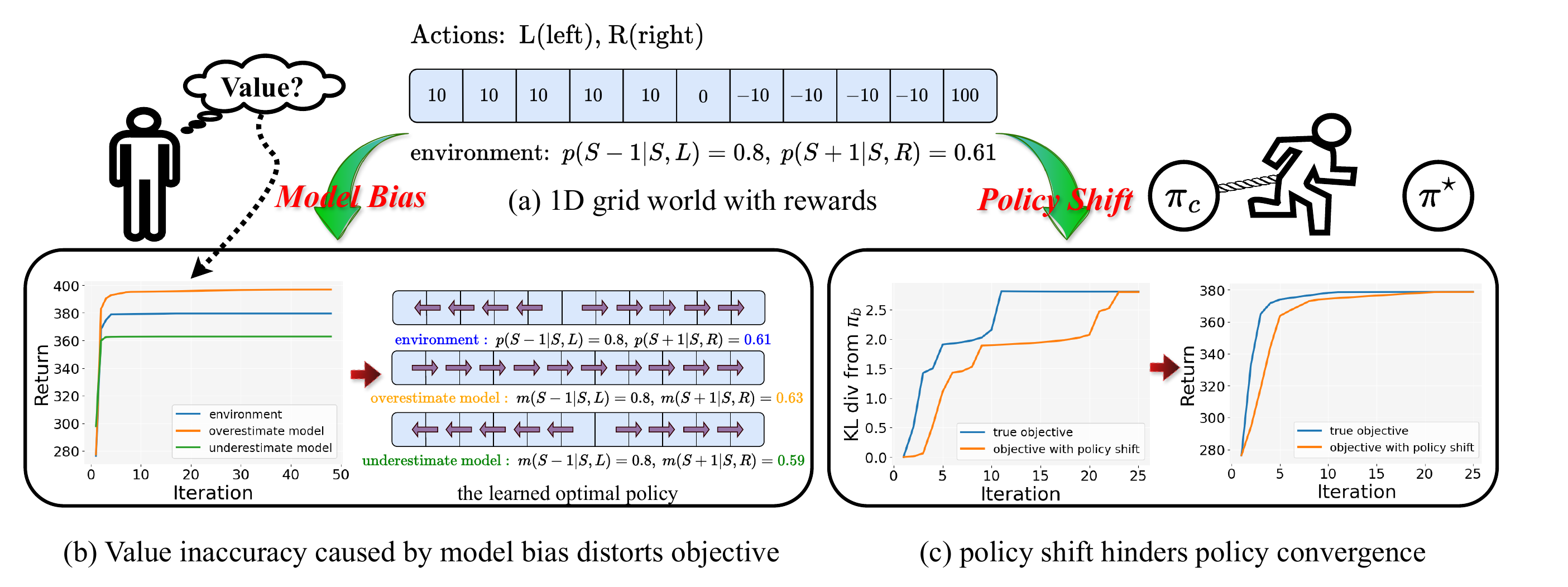} 
\caption{The negative impacts of model bias and policy shift. 
(a) In the 1D grid world, the agent can take two actions: move left (L) or move right (R) from the current state $S$ to the adjacent grid cells $S-1$ or $S+1$, with movement obstructed at the boundaries. The initial state is uniformly distributed, and the agent receives undiscounted rewards upon reaching specific states.
(b) The left panel illustrates value estimates during training and the final policy convergence with and without model bias. Even slight model bias can significantly distort value function estimates, leading the learned policies to deviate from the true optimal policy in the actual environment.
(c) The right panel showcases expected returns and the KL divergence between the learned policy and the behavior policy with and without policy shift. 
We find that policy shift implicitly slows policy convergence.
}
\label{fig: toy example}
\end{figure*}

In this paper, we demystify the distribution shift problem in model-based RL and study its fundamental properties. 
We first show that this problem arises from both model bias and policy shift, and further analyze their impacts on policy optimization.
Unlike the heuristic descriptions of model uncertainty, model bias fundamentally characterizes the inconsistency between the learned model and actual environment dynamics in predicting subsequent states.
This bias leads to inaccurate estimations of trajectory returns, resulting in distorted objectives. If the learned model overestimates or underestimates the probability of generating a trajectory, the resulting distribution shift during training consequently leads to overestimation or underestimation of trajectory returns.
Additionally, policy shift depicts the changes in state-action visitation distribution caused by the deviation between the training data collection policy and the learned policy. 
This distribution shift between training and deployment implicitly perturbs policy updates and thereby hinders learning.
Existing studies heuristically mitigate these issues with off-policy mechanisms and uncertainty penalties, lacking a consistent, theoretically grounded framework.

To address these challenges, we theoretically derive adjustments for model bias and policy shift by employing a unified probabilistic inference framework, and introduce a shifts-aware reward~(SAR), which integrates the vanilla reward with adjustments for model bias and policy shift.
The model bias adjustment penalizes the reward to mitigate issues of overestimation or underestimation, thereby refining value learning and alleviating distribution shift during training.
Meanwhile, the policy shift modification adjusts the reward to mitigate distribution shift between training and deployment,
stabilizing the policy updates and accelerating policy convergence.
Building on our theoretical framework, we develop Shifts-aware Model-based Offline Reinforcement Learning~(SAMBO-RL). This approach incorporates model bias adjustment for model-generated data and policy shift modification for training data. Furthermore, it learns a transition classifier and action classifiers to approximate the shifts-aware reward and then performs policy optimization based on this adjusted reward. Our contributions include:
\begin{itemize}
    \item 
    We demystify the distribution shift in model-based offline RL, \textit{i.e.}, model bias causes distribution inconsistency in the training data, ultimately distorting value estimation, while policy shift leads to distribution inconsistency between training and deployment, hindering policy convergence.
    \item We upgrade the vanilla reward into a shifts-aware reward that incorporates model bias adjustment and policy shift modification, effectively mitigating distribution shift.
    \item We devise SAMBO-RL, which trains a transition classifier and an action classifier to approximate the shifts-aware reward, effectively addressing both model bias and policy shift.
    \item We empirically show that shifts-aware reward effectively mitigates distribution shift, and SAMBO-RL demonstrates comparable performance across various benchmarks.
\end{itemize}


\section{Preliminaries}

\noindent
\textbf{MDP.} We focus on a Markov decision process (MDP) specified by the six-elements tuple $M=(\mathcal{S},\mathcal{A},p,r,\mu_0,\gamma)$, where $\mathcal{S}$ denotes the state space, $\mathcal{A}$ denotes the action
space, $p(s^{\prime}|s,a)$ is the environment transition dynamics, positive $r(s, a)>0$ is the reward function, $\mu_0(s)$ is the initial state distribution, and $\gamma \in (0,1)$ is the discount factor typically used for future reward weighting. Given an MDP, we define the state value function of the policy $\pi$: $V^\pi(s) = \mathbb{E}_{\pi,p}\left[ \sum_{t=0}^{\infty} \gamma^t r(s_t,a_t) | s_0=s\right]$ and the $Q$ function of the policy $\pi$: $Q^\pi(s,a) = \mathbb{E}_{\pi,p}\left[ \sum_{t=0}^{\infty} \gamma^t r(s_t,a_t) | s_0=s, a_0=a\right]$.

\quad
        
\noindent
\textbf{MBPO.}
Similar to previous works \cite{yu2020mopo,yu2021combo,rigter2022rambo,sun2023model} in offline RL, we focus on model-based policy optimization (MBPO) \cite{janner2019trust}, which uses an actor-critic RL algorithm.
Define $\mathcal{D}_{env}$ as the offline dataset collected by behavior policy $\pi_b$ in the real environment, and model dataset $\mathcal{D}_{m}$ as the buffer storing synthetic data. 
MBPO first performs $h$-step rollouts using the model $m(s^{\prime}|s,a)$ starting from state $s \in \mathcal{D}_{env}$ and adds the generated data to the buffer $\mathcal{D}_m$.
It then optimizes the policy using data sampled from $\mathcal{D}_{env} \cup \mathcal{D}_m$, where each datapoint is drawn from $\mathcal{D}_{env}$ with probability $f$ and $\mathcal{D}_m$ with probability $1 - f$. We denote the data used for training as $\mathcal{D}_{train}$.    

\quad  

\noindent
\textbf{Trajectory Formulation.}
To demystify the distribution shift, We consider the trajectory formulation of optimization objectives.
Let $p^{\pi}(\cdot)$ represent the probability distribution over trajectories $\mathcal{H}=\{\tau=(s_0,a_0,r_0,s_1,...)\}$ generated by executing policy $\pi$ under dynamics $p$. Then the generated probability is given by:
\begin{equation}\label{eq: true trajectory prob}
    p^{\pi}(\tau)=\mu_0(s_0)\prod \limits_{t=0}^{\infty}p(s_{t+1}|s_t,a_t)\pi(a_t|s_t).
\end{equation}
The training data $\mathcal{D}_{train}$ can be interpreted as generated by a data collection policy $\pi_c$ through rollouts in an unknown dynamics $q$. $\pi_c$ is inherently related to $\pi$, as the dataset $\mathcal{D}_{train}$ includes samples generated by the policy $\pi$. Consequently, the probability of observing trajectory $\tau$ in the training data is formulated as $q^{\pi_c}(\tau)$.
Then the RL objective and the practical training objective are 
\begin{align} \label{RL-object}
     \mathcal{J}_{M}(\pi)=\mathbb{E}_{\tau \sim p^{\pi}} \left[ R(\tau)\right] \text{and }
     \mathcal{J} (\pi) = \mathbb{E}_{\tau \sim q^{\pi_c}} [R(\tau)],
\end{align}
where $R(\tau)= \sum_{t=0}^{\infty} \gamma^t r(s_t,a_t)$ represents the discounted return of the whole trajectory $\tau=(s_0,a_0,s_1,a_1,...)$.


\section{Related Work}\label{appendix:related_work}

\textbf{Model-free Offline RL} encompasses two primary approaches:
policy constraints and value regularization, both incorporating conservatism~\cite{jin2021pessimism}. Policy constraints keep the learned policy close to the behavior policy, while value regularization adds conservative terms to the value optimization.
BEAR~\cite{kumar2019stabilizing} and BRAC~\cite{wu2019behavior} constrain the learned policy by minimizing different divergences. 
BCQ~\cite{fujimoto2019off} perturbs actions during learning to ensure they remain within the action space of the offline dataset by learning a generative model of the behavior policy.
SACS~\cite{mao2024offline} simultaneously performs out-of-distribution (OOD) state correction using a model that is not employed for data generation and OOD action suppression in offline RL.
CQL~\cite{kumar2020conservative} penalizes the Q-values of OOD samples by incorporating regularization into the value loss function. TD3+BC~\cite{fujimoto2021minimalist} introduces a behavior cloning regularization into the TD3~\cite{fujimoto2018addressing} objective. Similarly, EDAC~\cite{an2021uncertainty} and 
MCQ~\cite{lyu2022mildly} actively trains OOD actions by constructing pseudo-target values to alleviate over-pessimism.
DMG~\cite{mao2024doubly} leverages mild action generalization and generalization propagation to enhance generalization in offline RL.
SEABO assigns larger rewards to transitions closer to their nearest expert demonstrations, alleviating offline RL’s heavy reliance on the offline data annotated with reward labels.
A2PR~\cite{liu2024adaptive} selects high-advantage actions as an imitation learning regularizer to mitigate performance degradation from excessive conservatism.

\quad
        
\noindent
\textbf{Model-based Offline RL} approximates the environment using learned models and performs conservative policy optimization~\cite{lu2021revisiting}, achieving high data efficiency~\cite{li2024settling}. MOPO~\cite{yu2020mopo} incorporates conservatism by penalizing rewards based on the uncertainty of model predictions.
MOReL~\cite{kidambi2020morel} introduces a pessimistic MDP to penalize rewards of state-action pairs in unexplored regions. COMBO~\cite{yu2021combo} extends CQL to the model-based setting, regularizing the value function on OOD samples.
RAMBO~\cite{rigter2022rambo} adopts an adversarial approach, training the policy and model within a robust framework while ensuring accurate transition predictions of the model. 
CBOP~\cite{jeong2022conservative} adaptively weights multi-step returns in a model-based value expansion framework~\cite{feinberg2018model}.
CABI~\cite{lyu2022double} improves the reliability of synthesized data by training bidirectional dynamics models and trusting only the samples where the forward and backward models are consistent.
C-LAP~\cite{alles2024constrained} learns a generative model of the joint distribution of observations and actions, and imposes an implicit constraint on the generated actions to enable constrained policy learning.
Lastly, MOBILE~\cite{sun2023model} introduces penalties during value learning by leveraging the uncertainty of the Bellman Q-function estimates derived from ensemble models, achieving SOTA performance among these model-based and model-free approaches.

\quad
        
\noindent
\textbf{Cross-domain policy transfer} aims to transfer a policy learned in a source environment to a target environment with different dynamics.
This is similar to model-based approaches, which learn a model to approximate the environment dynamics and derive high-performing policies in the real environment by training within the learned model.
GARAT~\cite{desai2020imitation} learns an adversarial imitation-from-observation policy by discriminating between generated actions and target environment actions. 
DARC~\cite{eysenbach2020off} addresses cross-domain RL via reward correction to enable domain adaptation under changing environment dynamics.
SAIL~\cite{liu2019state} performs state alignment in cross-domain imitation learning to align state sequences across different domains.
Similarly, SRPO~\cite{xue2023state} introduces a regularization term based on the KL divergence between state distributions of different domains in the RL objective to improve data efficiency. 
OTDF~\cite{lyu2025cross} uses cross-domain offline datasets and mitigates dynamics mismatches via optimal transport, facilitating policy learning.


\section{Shifts-aware Policy Optimization}

We first formalize the distribution shift problem as shift weighting and further delve into its negative effects on policy optimization from theory and experiments.
To address these challenges, we employ probabilistic inference to derive a surrogate objective and formulate a shifts-aware reward.
Then we implement this theoretical framework within model-based offline RL and introduce SAMBO-RL, a practical algorithm to mitigate distribution shifts. Finally, we provide a theoretical analysis of SAR.

\subsection{Demystifying Distribution Shift}

We examine the distribution shift problem from an optimization perspective and identify its origins as model bias and policy shift. Then we explore how these factors affect accuracy in value estimation and induce instability in policy updates.

\textbf{Shift Weighting.}
The training objective $\mathcal{J} (\pi)$ in \eqref{RL-object} can be expressed as the trajectory formulation:
\begin{equation} \label{eq: training sampling formulation of true objective}
    \mathcal{J} (\pi)  
    = \mathbb{E}_{\tau \sim p^{\pi}(\tau)} \left[ \frac{q^{\pi_c}(\tau)}{p^{\pi}(\tau)} R(\tau) \right].
\end{equation}
Compared to $\mathcal{J}_{M}(\pi)$ in \eqref{RL-object}, it indicates that for a given trajectory $\tau$, the true objective under practical sampling distributions differs in shift weighting $\frac{q^{\pi_c}(\tau)}{p^{\pi}(\tau)}$ from the training objective,
which formalizes the distribution shift in trajectory formulation.
Substituting \eqref{eq: true trajectory prob} into the shift weighting $\frac{q^{\pi_c}(\tau)}{p^{\pi}(\tau)}$, we can attribute it to two primary components: model bias and policy shift. Then the shift weighting can be reformulated as the following: 
\begin{align} \label{D-S}
    \frac{q^{\pi_c}(\tau)}{p^{\pi}(\tau)}=
    \underbrace{\prod_t \frac{ q(s_{t+1} | s_t,a_t)}{ p(s_{t+1} | s_t,a_t)}}_{\text{model bias}}
    \underbrace{\prod_t \frac{ \pi_c(a_t | s_t)}{ \pi(a_t | s_t)}}_{\text{policy shift}}.
\end{align}
Model bias illustrates the discrepancy between the dynamics used for training and those of the actual environment, while policy shift represents the divergence between the policies used for sampling training data and those being learned. Both components ultimately lead to the distribution shift issue.

\textbf{Negative Impact of Shift Weighting.}
We analyze the impact of shift weighting from model bias and policy shift.

(\textbf{i}) To illustrate the effect of model bias,
we examine the synthetic data generation scenario in model-based approaches, where the training policy $\pi$ operates within model dynamics $m$ (\textit{i.e.} $q=m$), which diverges from the environment dynamics $p$. In this context, according to \eqref{D-S}, the training objective $\mathcal{J}(\pi)$ is:
\begin{align} \notag
      \mathcal{J}(\pi)  
     &=  \mathbb{E}_{\tau \sim p^{\pi}} \left[R(\tau) \prod \limits_{t=0}^{\infty} \frac{m(s_{t+1}|s_t,a_t)}{p(s_{t+1}|s_t,a_t)}\right] .
\end{align} 
Compared to the true objective, the compounded prediction bias of the model significantly affects value estimation and leads to a distorted objective. Specifically, when the model $m$ overestimates or underestimates the probability of generating trajectory $\tau$, it consequently overestimates or underestimates the expected return of this trajectory in the real environment.
The inconsistency between the model and the environment in predicting future dynamics leads to a mismatch in trajectory distributions during training. 
This distribution shift in training introduces bias into the optimization objective, thereby guiding the policy toward suboptimal or incorrect updates.

(\textbf{ii}) To delve deeper into the policy optimization process under policy shift, 
We consider a typical offline reinforcement learning scenario, in which policy optimization is constrained to the training dataset $\mathcal{D}_{train}$, and assume that the learned model perfectly fits the environment dynamics.
In this scenario, the training objective $\mathcal{J}(\pi)$ 
can be formulated as:
\begin{align} \notag
      \mathcal{J}(\pi) 
     &= \mathbb{E}_{(s,a) \sim d^{p,\pi}} \left[r(s,a) (d^{p,\pi_c}(s,a)/d^{p,\pi}(s,a))  \right] .
\end{align} 
Here, $d^{p,\pi}(s,a)$ is the state-action visitation distribution under dynamics $p$ and policy $\pi$ 
$$d^{p,\pi}(s,a) = (1-\gamma) \mathbb{E}_{s_0 \sim \mu_0, s_{t} \sim p, a_{t} \sim \pi} \left[ \sum_{t=0}^\infty \gamma^t \mathbb{I}(s_t = s, a_t = a) \right],$$ 
where $\mathbb{I}$ represents the indicator function.
Note that $\pi_c$ is related to $\pi$. 
Compared to the true objective, the policy shift implicitly forces policy $\pi$ to approach $\pi_c$, perturbing policy updates. 
Specifically, for a state-action pair $(s,a)$ associated with a high reward, ideal policy updates should increase $\pi(a|s)$. However, if the currently learned policy $\pi$ adequately approximates the optimal policy and satisfies $d^{p,\pi}(s,a) > d^{p,\pi_c}(s,a)$, the reward for $(s,a)$ can be underestimated. This situation could even prompt policy updates to decrease $\pi(a|s)$, contrary to the optimal direction. 
Similarly, for the state-action pair $(s,a)$ with a low reward, 
if the currently $\pi$ is reasonably good and leads to $d^{p,\pi}(s,a) < d^{p,\pi_c}(s,a)$, the reward can be overestimated, potentially resulting in policy updates to increase $\pi(a|s)$ that do not align with optimal policy directions.
The mismatch between the deployed policy and the training distribution can disrupt updates.

\textbf{Toy Examples.}
To empirically validate our analysis of model bias and policy shift, we use a 1D grid world example depicted in Fig.~\ref{fig: toy example} (a). 
In Fig.~\ref{fig: toy example} (b), we compare the training process with and without model bias and find the value function overestimation and underestimation under model bias. 
These results reveal that even slight deviations in model dynamics can significantly distort value function estimates, leading to a biased training objective. Consequently, the policies learned by the agents deviate from the optimal policy.
To assess the impact of the policy shift, we examine a model-free offline setting where offline data is pre-collected using a behavior policy $\pi_b$. Fig.~\ref{fig: toy example} (c) showcases that the KL divergence between the learned policy $\pi$ and behavior policy $\pi_b$ under policy shift is substantially reduced compared to the scenario without policy shift, sometimes even decreasing undesirably. 
This indicates that the policy shift is impeding the training process and substantially slowing down the policy convergence.

\subsection{Shifts-aware Reward via Probabilistic Inference} \label{4.2}

This subsection introduces a novel shifts-aware reward through a unified probabilistic inference framework, serving as a lower-bound surrogate for the true objective. 
We examine how this modified reward contributes to refining the value function and
constraining unstable policy updates, thereby mitigating the distribution shift in both training and real-world deployment.

\textbf{Shifts-aware Reward.}
We introduce the shifts-aware reward $\tilde{r}(s_t,a_t,s_{t+1})$ formulated within trajectories. It comprises the vanilla reward, adjustments for model bias
and modifications for policy shift, expressed as the following equation:
\begin{equation}
    \begin{split}
	\tilde{r}(s_t,a_t,s_{t+1})&=\log r(s_t,a_t) \\
        &+\frac{1} {(1-\gamma)\gamma^t} \left[ \log \frac{p(s_{t+1} | s_t,a_t)}{q(s_{t+1} | s_t,a_t)}
         + \log\frac{\pi(a_t | s_t)}{\pi_c(a_t | s_t)} \right].
    \label{reward_shift_theory}
    \end{split}
\end{equation}
This modified reward forms the basis of our surrogate objective function during training, which is defined as:
\begin{align}
    \mathcal{L}(\pi)=\mathbb{E}_{q^{\pi_c}(\tau)} \left[ \sum_{t=0}^{\infty} \gamma^t \tilde{r}(s_t,a_t,s_{t+1})\right]. 
    \label{RL_objective_PI}
\end{align}
Additionally, we establish that the surrogate objective $\mathcal{L}(\pi)$ serves as a lower bound for the true objective.
\begin{theorem}
    Let $\mathcal{L}(\pi)$ be the training objective defined in \eqref{RL_objective_PI} and $\tilde{r}$ denote the shifts-aware reward~\eqref{reward_shift_theory}. Then,
    $$
    \log \left[  \mathcal{J}_{M}(\pi) \right] \ge (1-\gamma)\mathcal{L}(\pi),\quad \forall\ \pi.
    $$
\end{theorem}

The theorem ensures that $\mathcal{L}(\pi)$ is a rational surrogate objective for $\mathcal{J}_{M}(\pi)$. Optimization objective based on this shifts-aware reward provides a methodological foundation for integrating shift adjustments seamlessly to uniformly address distribution shift caused by model bias and policy shift.

\textbf{Model Bias Adjustment.}
We now examine how this adjustment refines the value estimation and addresses the challenge of model bias.
By rolling out the learned policy $\pi$ in model $m$, we obtain a model-generated dataset $\mathcal{D}_m$. 
Then, when training on dataset $\mathcal{D}_{train}$, the modified objective restricted to $\mathcal{D}_m$ is computed as:
\begin{equation} 
\begin{aligned}
    \tilde{J}(\pi) &= \mathbb{E}_{s \sim \mathcal{D}_m, a\sim \pi(\cdot|s), s^{\prime} \sim \mathcal{D}_m} \left[ \tilde{r}(s,a,s^{\prime}) \right] \\
    &= \mathbb{E}_{(s,a) \sim \mathcal{D}_m} \left[ \log r(s,a) - \alpha D_{\operatorname{KL}}(m(\cdot | s,a) \| p(\cdot | s,a)) \right].
    \nonumber
\end{aligned}
\end{equation}
This modified reward consists of the logarithmic vanilla reward penalized by the KL divergence between the model and environment dynamics. 
This adjustment reduces value function misestimation by accounting for the accuracy of the learned model. By emphasizing data from regions with reliable model predictions, it mitigates the distribution shift in training data caused by model bias.

\textbf{Policy Shift Modification.}
To delve into the positive effect of policy shift modification, we consider the modified training objective in the training dataset $\mathcal{D}_{train}$ restricted to $\mathcal{D}_{env}$, in which case the dynamic $q$ satisfies $q=p$ and the data
collection policy satisfies $\pi_c=\pi_b$ in equation \eqref{D-S}. Then the modified objective is
\begin{equation}
\begin{aligned}
     \tilde{J}(\pi)& = \mathbb{E}_{(s,a,s^{\prime}) \sim \mathcal{D}_{env}} \left[ \tilde{r}(s,a,s^{\prime}) \right] \\
     &= \mathbb{E}_{s \sim \mathcal{D}_{env}}  \left[ \mathbb{E}_{a\sim \pi_b(\cdot|s)} \left[ \log r(s,a) \right] - \beta D_{\operatorname{KL}}(\pi_b(\cdot | s)||\pi(\cdot | s)) \right].
        \nonumber
\end{aligned}
\end{equation}
By introducing a penalty to the vanilla reward based on the KL divergence between offline data collection policy $\pi_b$ and testing policy $\pi$, this modification prevents $\pi$ from making overly aggressive updates, thus stabilizing the learning process. It reduces the bias between the offline data and the testing policy, helping to mitigate the distribution shift caused by policy shift.

\subsection{Shifts-aware Model-based Offline RL}
We apply the shifts-aware reward in model-based offline RL and develop the algorithm SAMBO-RL.

To adapt the shifts-aware reward for practical implementation, where training data is typically stored in transition format rather than trajectories, we introduce two weighting factors, $\alpha$ and $\beta$. These factors replace respective coefficients in reward function \eqref{reward_shift_theory}:
\begin{equation}\label{reward_shift_practical in sambo}
    \tilde{r}(s,a,s^{\prime})=\log r(s,a) + \alpha \log \frac{p(s^{\prime} | s,a)}{q(s^{\prime} | s,a)}+\beta \log\frac{\pi(a | s)}{\pi_c(a | s)}.
\end{equation}
These weighting factors ensure the reward and adjustments are properly balanced in transition-based learning scenarios, which is empirically validated in the following experiments. 
Moreover, as discussed in Section~\ref{4.2}, for data in $\mathcal{D}_{env}$, the dynamics satisfy $q=p$ and the policy satisfies $\pi_c=\pi_b$. For data in $\mathcal{D}_{m}$, we have $q=m$, and the data-collection policy can be approximated by $\pi$, \textit{i.e.}, $\pi_c=\pi$. As discussed in detail in Appendix~\ref{Classifiers Training}, we have
\begin{align}
    \tilde{r}(s,a,s^{\prime}) =
    \begin{cases}
        \log r(s,a) + \beta \log\frac{\pi(a | s)}{\pi_b(a | s)}, & \forall (s,a,s^{\prime}) \in \mathcal{D}_{env} \\
        \log r(s,a) + \alpha \log \frac{p(s^{\prime} | s,a)}{m(s^{\prime} | s,a)}, & \forall (s,a,s^{\prime}) \in \mathcal{D}_{m} 
    \end{cases}
    \nonumber
\end{align}

Furthermore, since the environment dynamics $p$ and the policy $\pi_b$ are typically unknown and cannot be exactly computed, we derive classifiers to estimate the shifts-aware reward \cite{eysenbach2020off,eysenbach2022mismatched,liu2021regret}. Details are shown in Appendix \ref{Appendix: Classifiers Estimation Derivations}. Specifically, these classifiers estimate $ \log \frac{p(s^{\prime} | s,a)}{m(s^{\prime} | s,a)}$ and $\log\frac{\pi(a | s)}{\pi_b(a | s)}$ respectively. The transition classifier $C_{\phi}(s,a,s^{\prime})$ estimates the probability that the transition $(s,a,s^{\prime})$ comes from the environment, while the action classifier $C_{\psi}(s,a)$ estimate the probability that a given state-action pair $(s,a)$ is generated by the current policy $\pi$.
We train these classifiers by minimizing the cross-entropy loss:
\begin{equation} \label{eq: classifiers loss}
    \mathcal{L}(\varphi) = - \mathbb{E}_{\mathcal{D}_1} [\log C_\varphi] - \mathbb{E}_{\mathcal{D}_2} [\log (1 - C_\varphi)] .
\end{equation}
For the transition classifier $C_\varphi = C_{\phi}(s,a,s^{\prime})$, $\mathcal{D}_1 = \mathcal{D}_{env}$ and $\mathcal{D}_2 = \mathcal{D}_m$.
For the action classifier $C_\varphi = C_{\psi}(s,a)$, $\mathcal{D}_1 = \mathcal{D}_{\pi}$ is the dataset collected by executing the current policy $\pi$, and $\mathcal{D}_2 = \mathcal{D}_{env}$.
Based on these classifiers, we approximate the shift-aware reward by
\begin{align}\label{eq: pratical_reward_shift}
    \tilde{r}(s,a,s^{\prime}) =
    \begin{cases}
        \log r(s,a) +  \beta \log\frac{C_\psi(s,a)}{1-C_\psi(s,a)}, & \forall (s,a,s^{\prime}) \in \mathcal{D}_{env} \\
        \log r(s,a) + \alpha \log\frac{C_\phi(s,a,s^{\prime})}{1-C_\phi(s,a,s^{\prime})}, & \forall (s,a,s^{\prime}) \in \mathcal{D}_{m} 
    \end{cases}
\end{align}
The procedure of SAMBO-RL is outlined in Algorithm \ref{sambo in mainbody}, with implementation details provided in Appendix \ref{Appendix: Implementation Details}.

\begin{algorithm}[tb]
\caption{SAMBO-RL}
\label{sambo in mainbody}
\begin{algorithmic}[1] 
\Require Offline dataset $\mathcal{D}_{env}$, learned dyanmics models $\{m_{\theta}^i\}_{i=1}^N$, initialized policy $\pi$, hyperparameters $\alpha$ and $\beta$.
\State  Train the probabilistic dynamics model $m_{\theta}(s^{\prime},r | s,a)= \mathcal{N}(\mu_{\theta}(s,a) , \Sigma_{\theta}(s,a))$ on $\mathcal{D}_{env}$.
\State Initialize the model dataset buffer $\mathcal{D}_{m} \leftarrow \varnothing$. 
    \For{$i = 1,2,..., N_{iter}$}
        \State Initialize the policy dataset buffer $\mathcal{D}_{\pi} \leftarrow \varnothing$. 
        \State Generate synthetic rollouts by model $m_{\theta}$. Add transition data in these rollouts to $\mathcal{D}_{m}$ and $\mathcal{D}_{\pi}$.
        \State Update classifiers $C_{\phi}$ and $C_{\psi}$ according to Eq.~\eqref{eq: classifiers loss}.
        \State Compute shifts-aware reward according to Eq.~\eqref{eq: pratical_reward_shift}.
        \State Run SAC~\cite{haarnoja2018soft} with shifts-aware reward to update policy $\pi$.     
    \EndFor
\end{algorithmic}
\end{algorithm}

\subsection{Theoretical Analysis}

We analyze the theoretical properties of SAR, with detailed proofs provided in Appendix~\ref{Appendix:Theoretical Analysis}.
We first introduce \(\xi\)-uncertainty quantifier, which characterizes the suboptimality of the derived policy.

\begin{definition}
\label{Def: xi-Uncertainty Quantifier}
(\(\xi\)-Uncertainty Quantifier \cite{jin2021pessimism}).  
The set of penalization \(\{\Gamma_h\}_{h \in [H]}\) forms a \(\xi\)-uncertainty quantifier if it holds with probability at least \(1 - \xi\) that 
\[
\left| \widehat{\mathbb{B}} \widehat{V}_{t+1}(s, a) - \mathbb{B} \widehat{V}_{t+1}(s, a) \right| \leq \Gamma_t(s, a),   \quad  \forall(s,a)\in \mathcal{S} \times \mathcal{A}
\]
where \(\widehat{V}_{t+1}\) is an estimated value function at step \(t+1\) and \(\widehat{\mathbb{B}}\) is an empirical Bellman operator estimating the Bellman operator \(\mathbb{B}\). 
\end{definition}

We can reformulate \eqref{reward_shift_practical in sambo} in a state-action style:
\begin{equation}\label{sar(s,a)}
\begin{aligned}
    \tilde{r}(s,a)&= \mathbb{E}_{ s^{\prime}\sim q(\cdot|s)} \left[ \tilde{r}(s,a,s^{\prime}) \right] \\
    & = 
     \log r(s,a)- \alpha D_{\operatorname{KL}}(q(\cdot | s,a) \| p(\cdot | s,a))- \beta \log \frac{\pi_c(a|s)}{\pi(a|s)} .
     \nonumber
\end{aligned}
\end{equation}

Since the model cannot perfectly fit the environment dynamics \( p \), we assume that the KL divergence satisfies
$
D_{\operatorname{KL}}(m(\cdot \mid s, a) \,\|\, p(\cdot \mid s, a)) \geq \epsilon_{KL} >0
$. Under this KL divergence assumption, we prove that the penalty of SAR $\tilde{r}(s,a)$ forms a \(\xi\)-uncertainty quantifier.


\begin{theorem}
\label{TH: SAR is xi-uncertainty quantifier}
Under the assumption that $
D_{\operatorname{KL}}(m(\cdot \mid s, a) \,\|\, p(\cdot \mid s, a)) \geq \epsilon_{KL} >0
$, there exist $\alpha>0$ and $\beta>0$ such that the penalty of SAR forms a \(\xi\)-uncertainty quantifier.
\end{theorem}

Based on this theorem, we derive the following value bound from the perspective of pessimistic value iteration (PEVI) \cite{jin2021pessimism}, which demonstrates that SAR is an effective corrective reward.

\begin{theorem}\label{TH: bound}
Let $\pi^*$ be the optimal policy in the environment, $\pi^{*}_{G}$ be the optimal policy in MDP $G=(\mathcal{S},\mathcal{A},p,\log r,\mu_0,\gamma)$, $V_{G}^{\pi}(s)$ the value function in $G$ and $
\epsilon_{s} \geq D_{\operatorname{KL}}(m(\cdot \mid s, a) \,\|\, p(\cdot \mid s, a)) \geq \epsilon_{KL} >0
$.
Then the derived policy \(\widehat{\pi}\) according to SAR satisfies 
\begin{equation}
\left| V^{\pi^*}(s_1) - V^{\widehat{\pi}}(s_1) \right| 
\leq 2 \alpha H \epsilon_s+\epsilon(G;s_1),
\nonumber
\end{equation}
with probability at least \(1 - \xi\) for all \(s \in \mathcal{S}\). 
Here,
$$\epsilon(G;s)=|V^{\widehat{\pi}}(s) + V_{G}^{\pi^{*}_{G}}(s) -V^{\pi^*}(s) -V_{G}^{\widehat{\pi}}(s)  |,$$
which quantify the value error between different MDPs and $\epsilon(M;s)=0, \forall s \in \mathcal{S}$.
\end{theorem}

\section{Experiments}

In our experiments, we focus on three objectives: (1) validating the effectiveness of shifts-aware reward, (2) comparing the performance of SAMBO with previous model-free and model-based methods, and (3) analyzing the contribution of each component in SAMBO.

We evaluate our algorithm using standard offline RL benchmarks in Mujoco environments and conduct an ablation study to assess its performance.  
Our implementation is based on the OfflineRL-Kit library\footnote{https://github.com/yihaosun1124/OfflineRL-Kit}, a comprehensive and high-performance library for implementing offline RL algorithms. The basic parameters of our algorithm are consistent with the settings in this library.

\begin{table*}[t]
    \centering
    \caption{Results on the D4RL Gym benchmark. 
    MOPO are obtained on the ``v0'' datasets; MOPO* are obtained from experiments in the OfflineRL-Kit library on the ``v2'' datasets.
    The numbers reported for SAMBO are the normalized scores averaged over the final iteration of training across 4 seeds, with $\pm$ standard deviation.
    The top three scores are highlighted in bold. 
    } 
    \vskip 0.1in
    \resizebox{\textwidth}{!}{
    \begin{tabular}{l|c c c c|c c c c c c|c}
        \toprule
        \textbf{Task Name} & \textbf{BC} & \textbf{CQL} & \textbf{TD3+BC} & \textbf{EDAC} & \textbf{MOPO} & \textbf{MOPO*} & \textbf{COMBO} & \textbf{TT} & \textbf{RAMBO} & \textbf{MOBILE(SOTA)} &\textbf{SAMBO (Ours)} \\
        \midrule
        halfcheetah-random & 2.2 & 31.3 & 11.0 & 28.4 & 35.4 & 38.5 & 38.8 & 6.1 & \textbf{39.5} & \textbf{39.3} & \textbf{39.7$\pm$2.0} \\
        hopper-random & 3.7 & 5.3 & 8.5 & 25.3 & 11.7 & \textbf{31.7} & 17.9 & 6.9 & 25.4 & \textbf{31.9} & \textbf{32.4$\pm$0.5} \\
        walker2d-random & 1.3 & 5.4 & 1.6 & \textbf{16.6} & 13.6 & 7.4 & 7.0 & 5.9 & 0.0 & \textbf{17.9} & \textbf{8.1 $\pm$ 7.8}\\
        \midrule
        halfcheetah-medium & 43.2 & 46.9 & 48.3 & 65.9 & 42.3 & 72.4 & 54.2 & 46.9 & \textbf{77.9} & \textbf{74.6} & \textbf{72.5$\pm$3.8} \\
        hopper-medium & 54.1 & 61.9 & 59.3 & \textbf{101.6} & 28.0 & 62.8 & 97.2 & 67.4 & 87.0 & \textbf{106.6} & \textbf{99.7$\pm$0.3} \\
        walker2d-medium & 70.9 & 79.5 & 83.7 & \textbf{92.5} & 17.8 & 84.1 & 81.9 & 81.3 & \textbf{84.9} & \textbf{87.7} & 78.1$\pm$2.3  \\
        \midrule
        halfcheetah-medium-replay & 37.6 & 45.3 & 44.6 & 61.3 & 53.1 & \textbf{72.1} & 55.1 & 44.1 & 68.7 & \textbf{71.7} & \textbf{70.6$\pm$1.6} \\
        hopper-medium-replay & 16.6 & 86.3 & 60.9 & \textbf{101.0} & 67.5 & 92.7 & 89.5 & 99.4 & 99.5 & \textbf{103.9} & \textbf{101.5$\pm$1.5} \\
        walker2d-medium-replay & 20.3 & 76.8 & 81.8 & 87.1 & 39.0 & 85.9 & 56.0 & 82.6 & \textbf{89.2} & \textbf{89.9} &  \textbf{92.3$\pm$ 1.5} \\
        \midrule
        halfcheetah-medium-expert & 44.0 & 95.0 & 90.7 & \textbf{106.3} & 63.3 & 83.6 & 90.0 & 95.0 & 95.4 & \textbf{108.2} & \textbf{ 97.8 $\pm$0.3 } \\
        hopper-medium-expert & 53.9 & 96.9 & 98.0 & 110.7 & 23.7 & 74.6 &\textbf{111.1}& 110.0 & 88.2 & \textbf{112.6} & \textbf{111.0 $\pm$1.2} \\
        walker2d-medium-expert & 90.1 & 109.1 & \textbf{110.1} & \textbf{114.7} & 44.6 & 108.2 & 103.3 & 101.9 & 56.7 & \textbf{115.2} & 106.1 $\pm$ 2.9 \\
        \midrule   
        \textbf{Average} & 36.5 & 61.6 & 58.2 & \textbf{76.0} & 36.7 & 70.3 & 66.8 & 62.3 & 67.7 & \textbf{80.0} &\textbf{75.8$\pm$ 2.2}\\
        \bottomrule
    \end{tabular}
    }
    \label{table: score in d4rl}
    \vskip 0.1in
\end{table*}

\begin{figure*}[h]
    \centering
    \begin{minipage}[t]{0.33\linewidth}
        \centering
        \includegraphics[width=1\columnwidth]{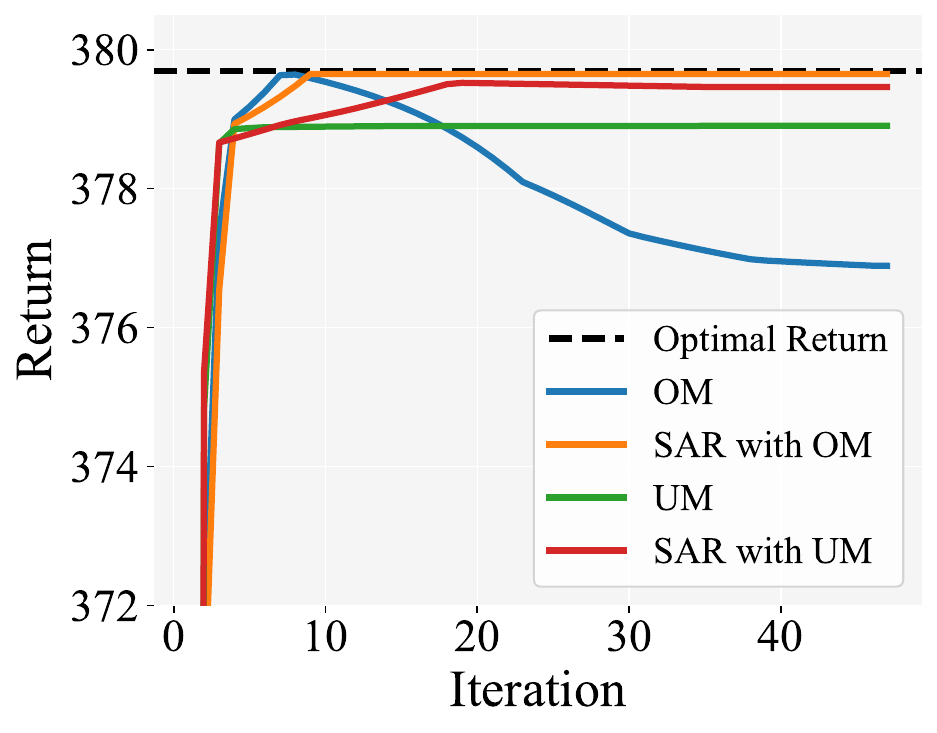}
    \end{minipage}
    \begin{minipage}[t]{0.33\linewidth}
        \centering
        \includegraphics[width=0.8\columnwidth]{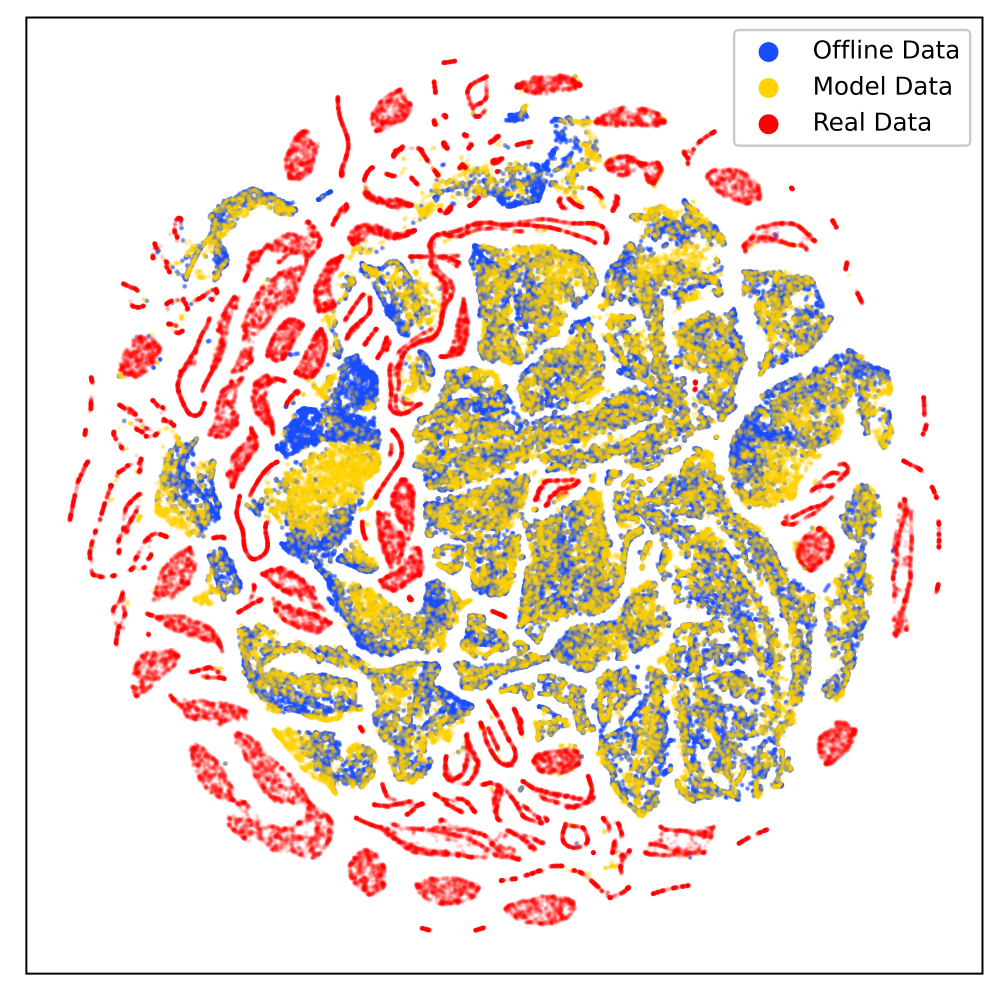}
    \end{minipage}%
    \begin{minipage}[t]{0.33\linewidth}
        \centering
        \includegraphics[width=0.8\columnwidth]{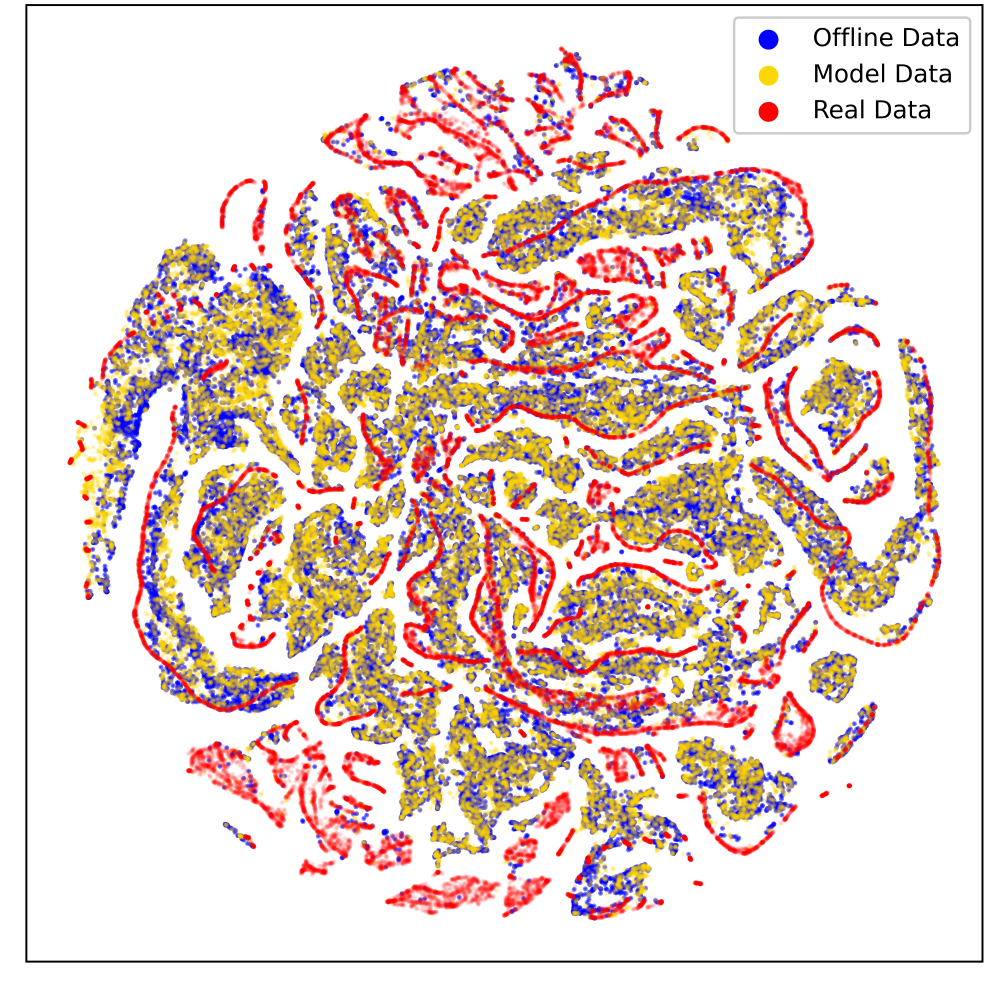}
    \end{minipage}
    \\
    \vskip 0.03in
    \begin{minipage}{0.33\linewidth}
        \centering
        \normalsize{(a) SAR in Toy Example} 
    \end{minipage}
    \begin{minipage}{0.33\linewidth}
        \centering
        \normalsize{(b) MOBILE} 
    \end{minipage}
    \begin{minipage}{0.33\linewidth}
        \centering
        \normalsize {(c) SAMBO}
    \end{minipage}
    \caption{Effectiveness of the shifts-aware reward. 
    (a) SAR is capable of achieving a near-optimal return.
    (b) and (c) The $(s, a, s^{\prime})$ distributions of offline RL algorithms result in the Hopper-M task in NeoRL benchmark. Model Data represents the distribution of data generated by executing the final learned policy in the dynamic model, while Real Data represents the distribution of data generated by executing the final learned policy in the real environment.}
    \label{fig: toy example with SAR}
    \vskip -0.1in
\end{figure*}

\subsection{Effectiveness of Shifts-aware Reward}
\textbf{Effectiveness in Toy Example.} 
To assess the efficacy of shifts-aware reward in different models, we examine its amendment to the training process. 
As illustrated in Fig.~\ref{fig: toy example with SAR} (a), 
we execute the policy gradient algorithm under four settings: using the vanilla reward with the overestimating model ({OM}) or underestimating model ({UM}), the SAR with the overestimating model ({SAR with OM}) or underestimating model ({SAR with UM}).
We then evaluate the expected return of the learned policy after each training update in the real environment.
The results show that while the {OM} initially achieves optimal returns, it rapidly diverges from the optimal policy. In contrast, {SAR with OM} effectively counters the performance degradation caused by model overestimation, sustaining optimal returns.
Similarly, {SAR with UM} mitigates the policy suboptimality due to model underestimation, achieving a return near the theoretical optimum. 
Both demonstrate the effectiveness.

\quad

\noindent
\textbf{Effectiveness in Complex Tasks.}
We also conducted experiments on complex tasks.
Fig.~\ref{fig: toy example with SAR} (c) and (d) shows a visual comparison in the NeoRL task (details in Sec.~\ref{section: neorl}). 
SAMBO effectively constrains model-generated data (i.e., Model Data) to align closely with the Offline Data by model bias adjustment, whereas the previous state-of-the-art (SOTA) method, MOBILE, imposes weaker constraints. 
This enables SAMBO to effectively address the distribution shift between offline data and model data during training, thereby allowing for more efficient and reliable model utilization.
Benefiting from the policy shift modification, the real-world data (i.e., Real Data) generated by SAMBO is well-aligned with both the Offline Data and the Model Data.
This enables SAMBO to align with the deployment distribution, effectively addressing the distribution shift between the training data and the data encountered during real deployment, facilitating an efficient connection from training to testing.
These ultimately result in an effective resolution to the distribution shift.

\begin{table*}[h]
\centering
\caption{Normalized average returns in NeoRL benchmark over 4 random seeds. The hightest scores are highlighted in bold.
}
\resizebox{0.75\textwidth}{!}{
\begin{tabular}{lccccccc}
\toprule
\textbf{Task Name} & \textbf{BC} & \textbf{CQL} & \textbf{TD3+BC} & \textbf{EDAC} & \textbf{MOPO} & \textbf{MOBILE}  & \textbf{SAMBO (Ours)}\\
\midrule
HalfCheetah-L & 29.1 & 38.2 & 30.0 & 31.3 & 40.1 &\textbf{54.8} & 38.4 $\pm$ 3.2\\
Hopper-L      & 15.1 & 16.0 & 15.8 & 18.3 & 6.2  &18.2 & \textbf{19.4 $\pm$ 1.3 }\\
Walker2d-L    & 28.5 & 44.7 & 43.0 & 40.2 & 11.6 & 23.7& \textbf{46.1 $\pm$ 5.7 }\\
\midrule
HalfCheetah-M & 49.0 & 54.6 & 52.3 & 54.9 & 62.3 & \textbf{79.0} & 62.5 $\pm$1.6 \\
Hopper-M      & 51.3 & 64.5 & \textbf{70.3} & 44.9 & 1.0  & 43.4& \textbf{67.6 $\pm$ 12.2 }\\
Walker2d-M    & 48.7 & 57.3 & 58.6 & 57.6 & 39.9 & 60.1& \textbf{63.4 $\pm$ 4.0 }\\
\midrule
HalfCheetah-H & 71.3 & 77.4 & 75.3 & \textbf{81.4} & 65.9 &  71.8& 71.6 $\pm$ 13.9\\
Hopper-H      & 43.1 & 76.6 & 75.3 & 52.5 & 11.5 & 42.3& \textbf{93.9 $\pm$ 15.2 }\\
Walker2d-H    & 72.6 & 75.3 & 69.6 & 75.5 & 18.0 & 71.9& \textbf{79.2 $\pm$ 0.6}\\
\midrule
\textbf{Average} & 45.4 & 56.1 & 54.5 & 50.7 & 28.5  & 51.7 & \textbf{60.2 $\pm$ 6.4}\\
\bottomrule
\end{tabular}
}
\label{table: score in neorl}
\end{table*}

\begin{figure*}[h]
 \vskip 0.1in
    \centering
    \begin{minipage}[t]{0.32\linewidth}
        \centering
        \includegraphics[width=0.9\columnwidth]{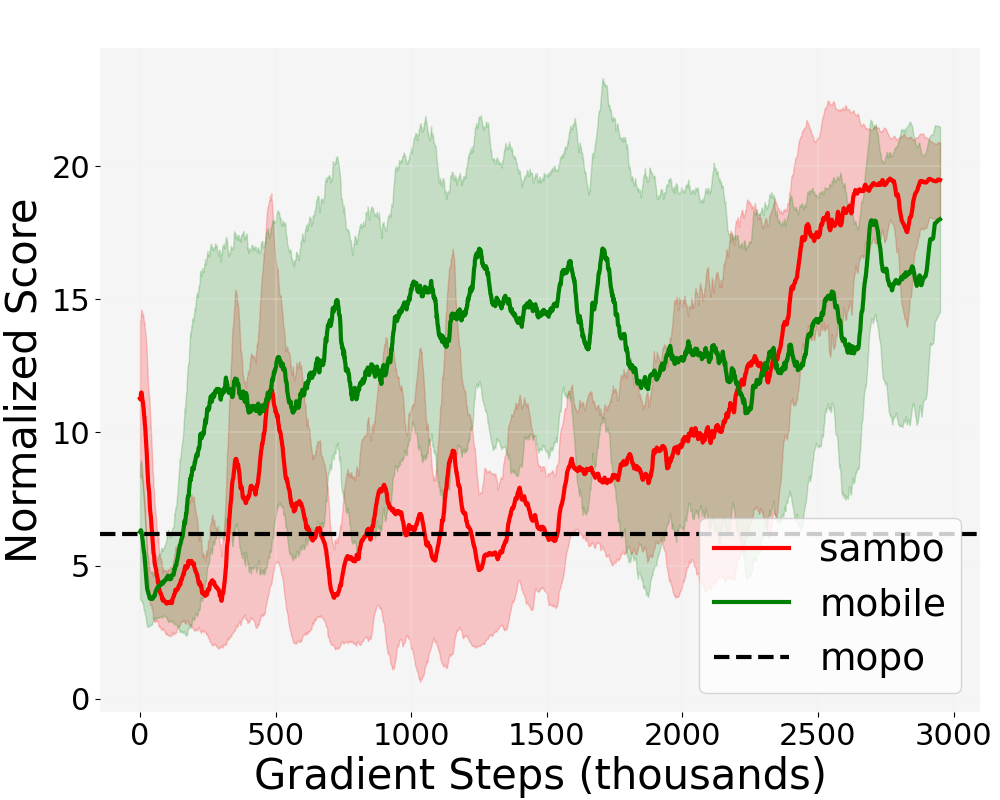}
    \end{minipage}
    \begin{minipage}[t]{0.32\linewidth}
        \centering
        \includegraphics[width=0.9\columnwidth]{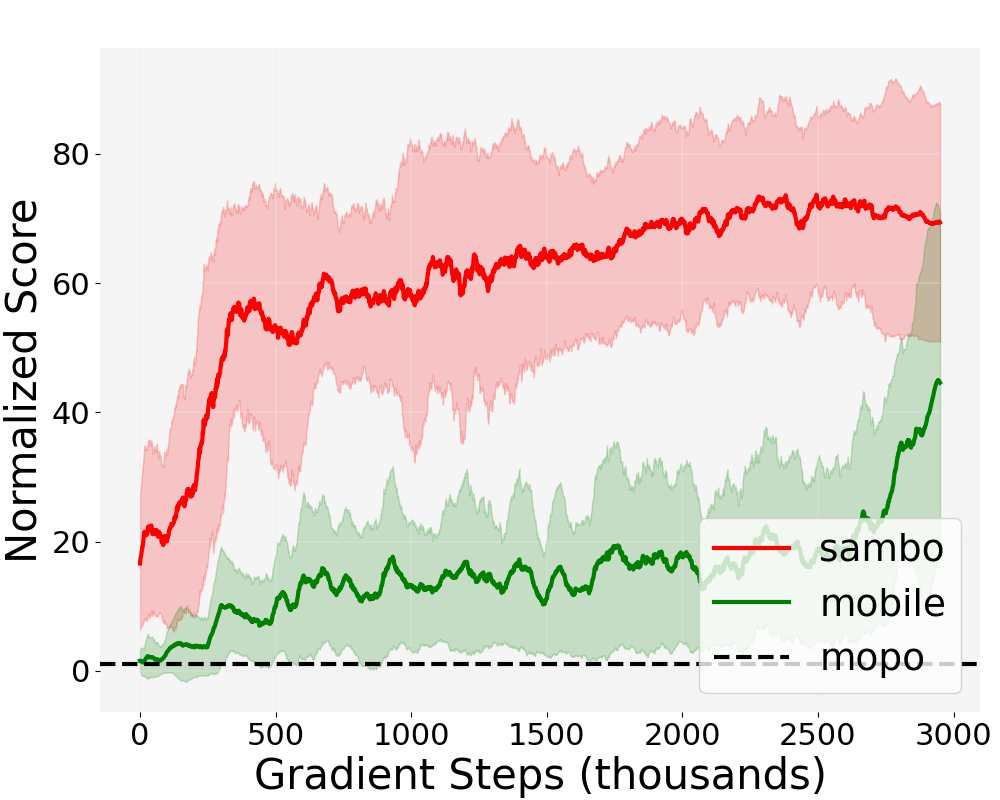}
    \end{minipage}
    \begin{minipage}[t]{0.32\linewidth}
        \centering
        \includegraphics[width=0.9\columnwidth]{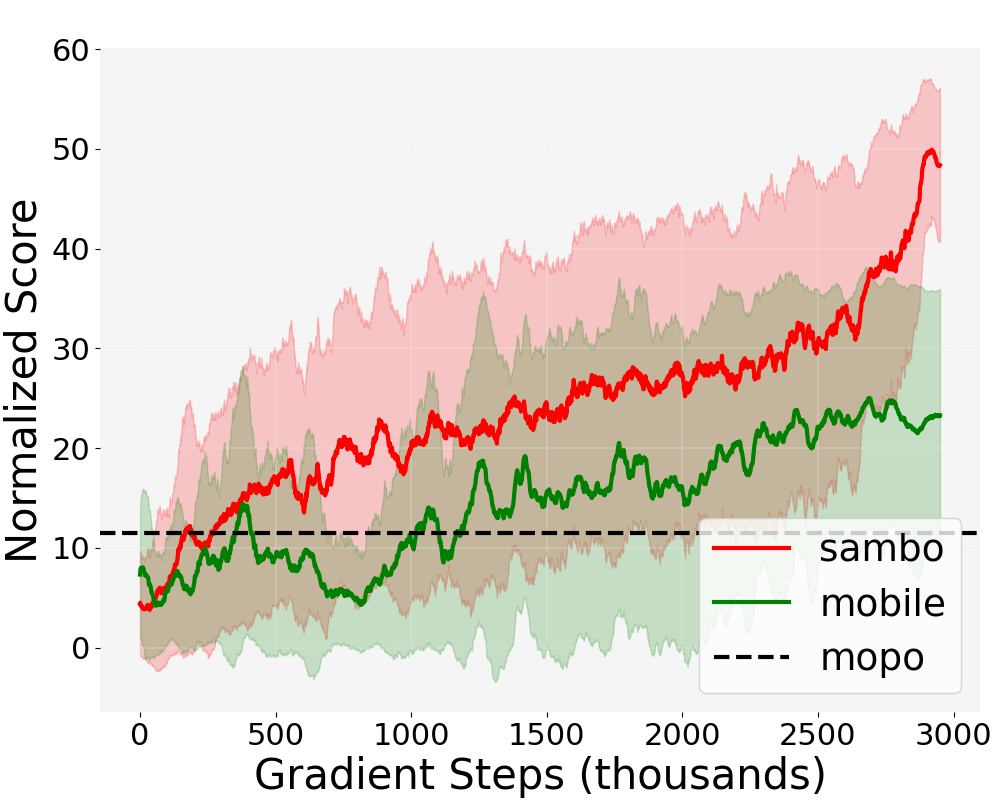}
    \end{minipage}
    \\
    \vskip 0.02in
    \begin{minipage}{0.32\linewidth}
        \centering
        {(a) Hopper-L} 
    \end{minipage}
    \begin{minipage}{0.32\linewidth}
        \centering
         {(b) Hopper-M}
    \end{minipage}
    \begin{minipage}{0.32\linewidth}
        \centering
         {(c) Hopper-H}
    \end{minipage}
    \caption{Performance comparison of Hopper tasks in NeoRL benchmark.}
    \label{fig: neorl performance compare}  
\end{figure*}

\subsection{Benchmark Results}

\subsubsection{D4RL} \label{D4RL experiment}

\quad

\noindent
\textbf{Datasets.}
In line with prior research, SAMBO is evaluated using the D4RL benchmark~\cite{fu2020d4rl} on the MuJoCo simulator~\cite{todorov2012mujoco}. The evaluation encompasses 12 datasets spanning three environments (halfcheetah, hopper, walker2d) and four dataset types (random, medium, medium-replay, medium-expert). 
To ensure consistency, we employ the ``v2'' versions of datasets for standardized evaluation.

\quad

\noindent
\textbf{Baselines.} We evaluate SAMBO against several baseline methods, including model-free methods such as BC~\cite{bain1995framework, ross2011reduction} which learns the behavior policy, CQL~\cite{kumar2020conservative} that penalizes Q-values on OOD samples, TD3+BC~\cite{fujimoto2021minimalist} that adopts a BC constraint during optimizing policy, and EDAC~\cite{an2021uncertainty} that penalizes the
Q-function based on uncertainty derived from ensemble networks, as well as model-based methods such as MOPO~\cite{yu2020mopo} that penalize rewards based on the uncertainty of the model
predictions, COMBO~\cite{yu2021combo} that extends the penalty of CQL in the model-based setting to regularize OOD samples, TT~\cite{janner2021offline} that employs a Transformer to model distributions and uses beam search to plan, RAMBO~\cite{rigter2022rambo} that trains the policy and model within a robust framework, and the prior SOTA algorithm MOBILE~\cite{sun2023model} that penalizes
value function during training by the uncertainty of Bellman Q-function estimates derived from ensemble models.


\noindent
\textbf{Comparison Results.} 
Results are presented in Table \ref{table: score in d4rl}. Each number represents the normalized score, calculated as $100\times$(score~$-$ random policy score)/ (expert policy score~$-$ random policy score)~\cite{fu2020d4rl}. SAMBO achieves strong performance on random datasets and demonstrates competitive performance across medium, medium-replay, and medium-expert datasets. Furthermore, SAMBO demonstrates highly competitive average performance, approaching that of the model-free ensemble method EDAC~\cite{an2021uncertainty}, and second only to the current SOTA model-based ensemble method MOBILE~\cite{sun2023model}. However, unlike these methods, SAMBO achieves this competitive performance without relying on value ensemble techniques.


\begin{figure*}[t]
    \centering
     \begin{minipage}[t]{0.32\linewidth}
        \centering
        \includegraphics[width=0.9\columnwidth]{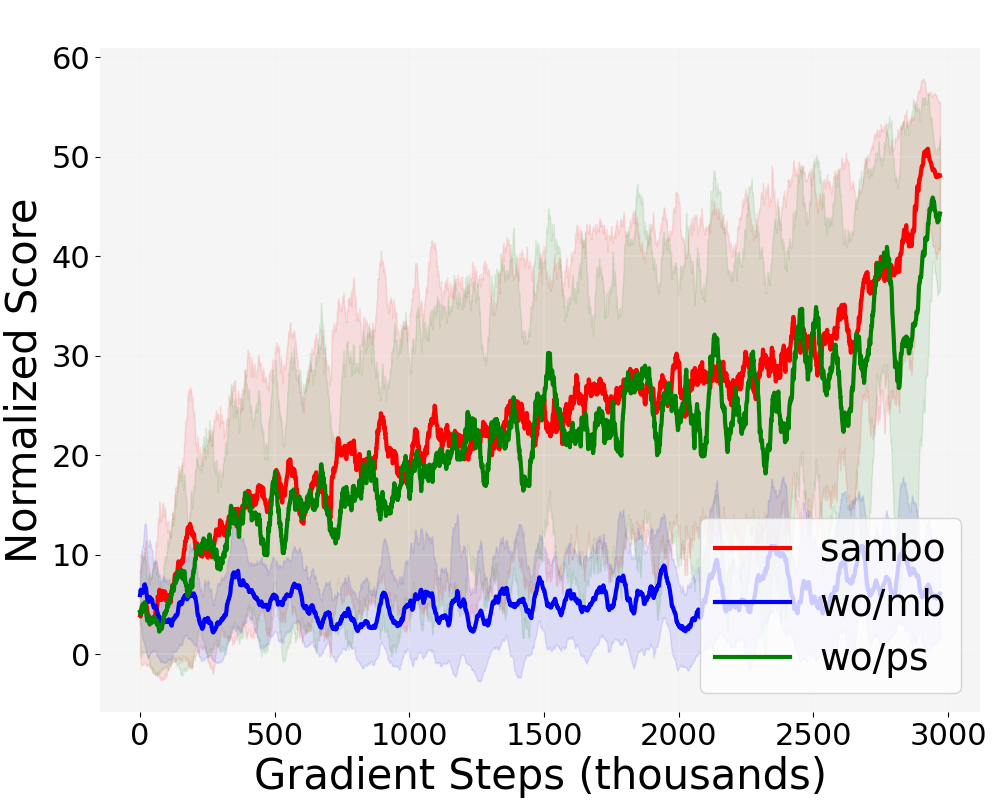}
    \end{minipage}
     \begin{minipage}[t]{0.32\linewidth}
        \centering
        \includegraphics[width=0.9\columnwidth]{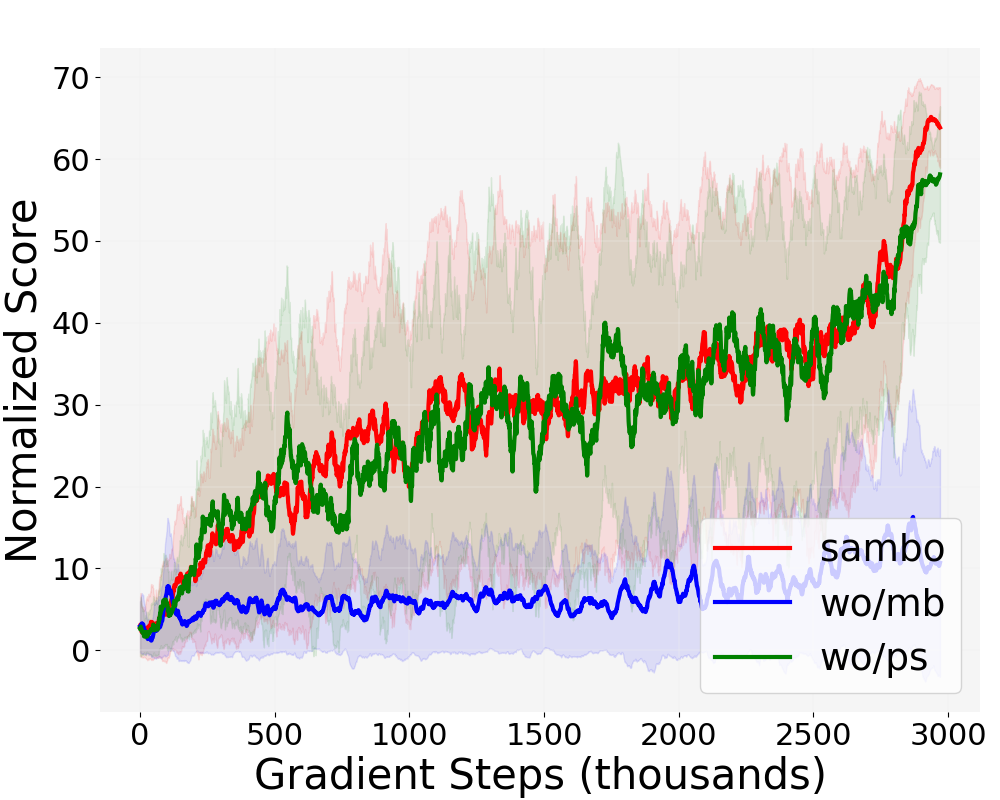}
    \end{minipage}
     \begin{minipage}[t]{0.32\linewidth}
        \centering
        \includegraphics[width=0.9\columnwidth]{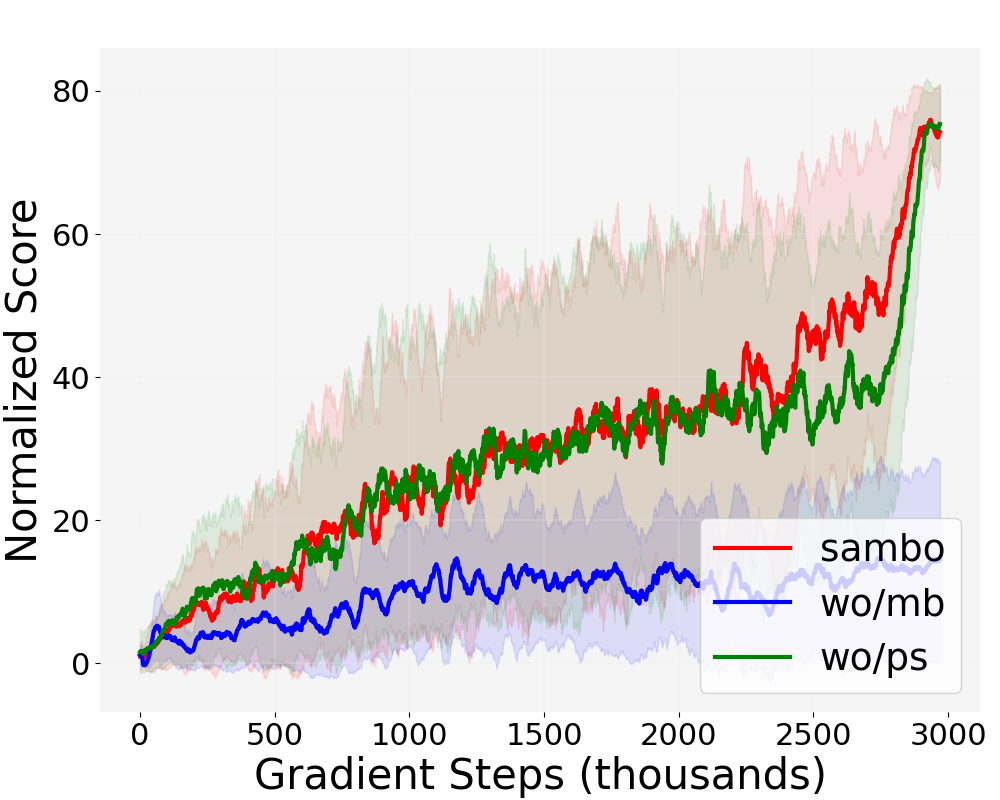}
    \end{minipage}
    \\
    \vskip 0.02in
    \begin{minipage}{0.32\linewidth}
        \centering
        \footnotesize{(a) Walker2d-L} 
    \end{minipage}
    \begin{minipage}{0.32\linewidth}
        \centering
        \footnotesize {(b) Walker2d-M}
    \end{minipage}
    \begin{minipage}{0.32\linewidth}
        \centering
        \footnotesize {(c) Walker2d-H}
    \end{minipage}
    \caption{Illustration of ablation experiments in NeoRL benchmark.
    }
    \label{fig: halfcheetah_random_ablation}
\end{figure*}

\subsubsection{NEORL}
\label{section: neorl}

\quad

\noindent
\textbf{Datasets.} NeoRL~\cite{qin2022neorl} is a benchmark carefully designed to emulate real-world scenarios by generating datasets through a more conservative policy, thereby closely aligning with real-world data-collection scenarios. The datasets, characterized by their narrow scope and limited coverage, present a substantial challenge for offline RL algorithms. In this work, we examine nine datasets across three environments (HalfCheetah-v3, Hopper-v3, Walker2d-v3) and three dataset quality levels (L, M, H), corresponding to low, medium, and high-quality datasets. NeoRL provides varying numbers of trajectories for each task (100, 1000, and 10000), and for consistency, we uniformly selected 1000 trajectories in all our experiments.


\noindent
\textbf{Baselines.} We compare SAMBO with baseline methods that align with the D4RL experiments in Section \ref{D4RL experiment}, excluding TT, COMBO, and RAMBO, as no results for these methods are reported in their original papers or the NeoRL paper, and determining suitable hyperparameters for them would be computationally expensive.

\quad

\noindent
\textbf{Comparison Results.} Results are presented in Table \ref{table: score in neorl} and Figure \ref{fig: neorl performance compare}. We compare the learning curves of model-based approaches in the NeoRL benchmark. Since MOPO does not have an official implementation on NeoRL, we report its score based on the original NeoRL paper. 
Compared to other model-based and model-free methods, including ensemble approaches, SAMBO achieves superior performance on the more challenging and realistic NeoRL benchmarks.
Moreover, SAMBO consistently outperforms Behavior Cloning (BC), while most baselines show degraded performance compared to BC on certain tasks.
In particular, SAMBO demonstrates exceptional performance improvements in the Walker2d and Hopper environments. 
These results highlight that SAMBO has strong potential for applications on challenging datasets. More experiments and discussions are provided in Appendix \ref{Appendix: Performance Compare}.



\subsection{Ablation and Tuning Studies}


Our shifts-aware reward integrates both the model bias adjustment and the policy shift modification. To elucidate the importance of these components, we perform ablation studies on the Walker2d task in NeoRL benchmark. As shown in Figure.~\ref{fig: halfcheetah_random_ablation},
we assess the performance under four configurations: SAMBO, SAMBO without model bias adjustment ({wo/mb}), SAMBO without policy shift modification ({wo/ps}), while keeping all other parameters consistent.

The ablations show that, 
The {wo/mb} employs policy shift modification to stabilize policy updates. However, it does not address model bias and leads to significant inaccuracies in value estimation, ultimately resulting in poor performance.
Conversely, the {wo/ps} ablation addresses model bias but neglects policy shift adjustments. While this reduces value misestimation, the algorithm remains unstable due to the induced distribution shift between training and deployment. 
Consequently, {wo/ps} exhibits lower stability, and its final performance is slightly worse than that of SAMBO.
In contrast, SAMBO incorporates both model bias adjustment and policy shift modification, facilitating stable updates and addressing value estimation challenges, thereby alleviating suboptimal performance.

\begin{figure}[t]
    \begin{minipage}[t]{0.48\linewidth}
        \centering
        \includegraphics[width=0.85\columnwidth]{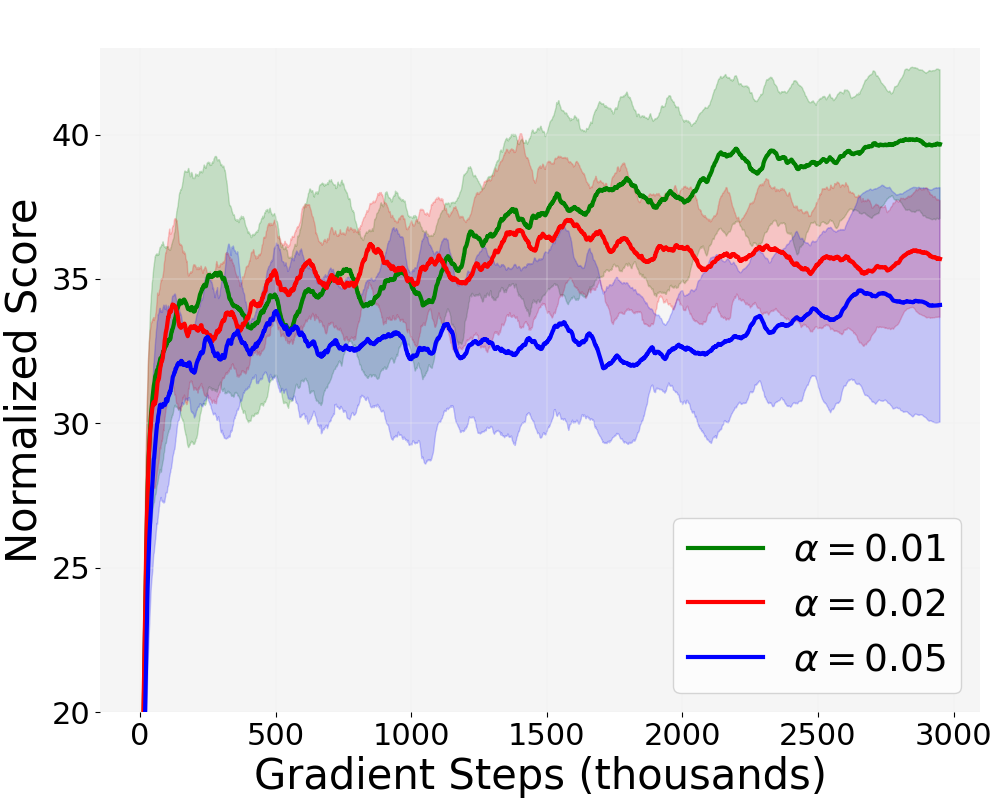}
    \end{minipage}%
    \begin{minipage}[t]{0.48\linewidth}
        \centering
        \includegraphics[width=0.85\columnwidth]{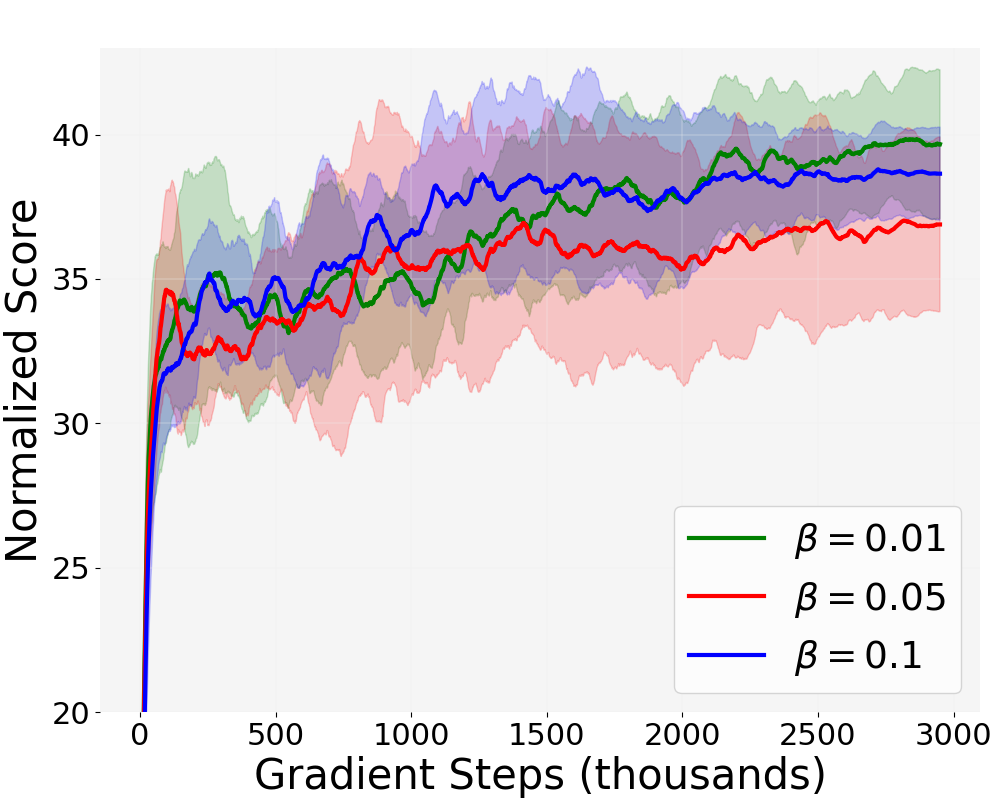}
    \end{minipage}
    \\
    \begin{minipage}{0.48\linewidth}
        \centering
        \footnotesize{(a) Tuning for $\alpha$} 
    \end{minipage}
    \begin{minipage}{0.48\linewidth}
        \centering
        \footnotesize {(b) Tuning for $\beta$}
    \end{minipage}
    \caption{ The tuning experiment was performed on halfchetah-random datasets in D4RL benchmark.}
    \label{fig: hyperparameter tuning}
\end{figure}

Furthermore, we performed parameter tuning experiments on $\alpha$ and $\beta$ to examine the robustness of our method with respect to parameter choices. As illustrated in Figure~\ref{fig: hyperparameter tuning}, we fix $\beta=0.01$ in example (a) and $\alpha=0.01$ in example (b). The results indicate that the performance of our algorithm remains generally stable and consistent within a certain range of parameter variations.


\section{Conclusion and Discussion}
\label{Conclusion}

In this paper, we analyze the distribution shift problem in model-based offline RL, attributing its root causes to model bias and policy shift. 
Model bias induces a distribution shift among training data originating from different sources,
which distorts value estimation, while policy shift results in a distribution shift between training and deployment, hindering policy convergence.
We introduce a shifts-aware policy optimization framework to uniformly address these challenges through shift-aware rewards~(SAR). 
Theoretical and empirical analyses have demonstrated that SAR serves as an efficient surrogate for the true rewards.
Building on SAR, we apply it within the model-based offline RL setting, proposing our SAMBO-RL algorithm.
Empirical results demonstrate that SAMBO-RL consistently delivers superior or competitive performance, especially on the more challenging NeoRL benchmarks, underscoring SAMBO's practical effectiveness and validating our theoretical analysis.
Our main limitation is classifier accuracy; imprecise classifiers struggle to estimate the shifts-aware reward accurately, leading to unstable training.
In future work, We aim to explore alternative methods that directly address distribution shifts using datasets, thereby eliminating the need for explicit classifier estimation.



\bibliographystyle{ACM-Reference-Format} 
\bibliography{AAMAS_2026_sample}


\newpage
\appendix
\section{Theoretical Analysis} \label{Appendix:Theoretical Analysis}

\subsection{The Proof of Theorem 1}
\setcounter{theorem}{0}
\begin{theorem}
    Let $\mathcal{L}(\pi)$ be the training objective defined in \eqref{RL_objective_PI} and $\tilde{r}$ denote the shifts-aware reward~\eqref{reward_shift_theory}. Then,
    $$
    \log \left[  \mathcal{J}_{M}(\pi) \right] \ge (1-\gamma)\mathcal{L}(\pi),\quad \forall\ \pi.
    $$
\end{theorem}

\begin{proof}
   The cumulative discounted reward $R(\tau)$ of trajectory $\tau=(s_0,a_0,r_0,s_1,...)$ can be written as:
    \begin{equation}
        R(\tau)=\sum_{t=0}^{\infty} \gamma^t r(s_t,a_t)=\frac{1}{1-\gamma} \sum_{t=0}^{\infty} (1-\gamma) \gamma^t r(s_t,a_t).
        \label{trajectory reward}
    \end{equation} 
    As discussed in the main text, the probability distribution over a trajectory $\tau$ can be formulated as follows:
    \begin{equation}
        p^{\pi}(\tau)=\mu_0(s_0)\prod \limits_{t=0}^{\infty}p(s_{t+1}|s_t,a_t)\pi(a_t|s_t)
        \label{trajectory fomulation p}
    \end{equation}
    \begin{equation}
        q^{\pi_c}(\tau)=\mu_0(s_0)\prod \limits_{t=0}^{\infty}q(s_{t+1}|s_t,a_t)\pi_c(a_t|s_t).
        \label{trajectory fomulation q}
    \end{equation}
    Then according to \eqref{trajectory reward}, \eqref{trajectory fomulation p} and \eqref{trajectory fomulation q} and Jensen's inequality, we have


    \begin{align*}
    &\log [ \mathcal{J}_{M}(\pi) ] \\
    &= \log \mathbb{E}_{p^{\pi}(\tau)} R(\tau) \\
    &= \log \mathbb{E}_{q^{\pi_c}(\tau)} [ R(\tau) \tfrac{p^{\pi}(\tau)}{q^{\pi_c}(\tau)} ]\\
    &= \log \mathbb{E}_{q^{\pi_c}(\tau)} \!\Big[ 
        R(\tau) 
        \tfrac{
            \mu(s_0)\prod_{t=0}^{\infty} p(s_{t+1} | s_t,a_t)\pi(a_t | s_t)
        }{
            \mu(s_0)\prod_{t=0}^{\infty} q(s_{t+1} | s_t,a_t)\pi_c(a_t | s_t)
        } 
    \Big]\\
    &\ge \mathbb{E}_{q^{\pi_c}(\tau)} \!\Big[ 
        \log R(\tau)
        + \sum_{t=0}^{\infty} (
            \log \tfrac{p(s_{t+1} | s_t,a_t)}{q(s_{t+1} | s_t,a_t)}
            + \log \tfrac{\pi(a_t | s_t)}{\pi_c(a_t | s_t)}
        )
    \Big]\\
    &\ge \mathbb{E}_{q^{\pi_c}(\tau)} \!\Big[
        \log \tfrac{1}{1-\gamma}
        + \sum_{t=0}^{\infty} (1-\gamma)\gamma^t \log r(s_t,a_t) \\
    &\quad
        + \sum_{t=0}^{\infty} (
            \log \tfrac{p(s_{t+1} | s_t,a_t)}{q(s_{t+1} | s_t,a_t)}
            + \log \tfrac{\pi(a_t | s_t)}{\pi_c(a_t | s_t)}
        )
    \Big]\\
    &\ge \mathbb{E}_{q^{\pi_c}(\tau)} \!\Big[
        \sum_{t=0}^{\infty} (
            (1-\gamma)\gamma^t \log r(s_t,a_t)
            + \log \tfrac{p(s_{t+1} | s_t,a_t)}{q(s_{t+1} | s_t,a_t)}
            + \log \tfrac{\pi(a_t | s_t)}{\pi_c(a_t | s_t)}
        )
    \Big]\\
    &= (1-\gamma) \mathbb{E}_{q^{\pi_c}(\tau)} \!\Big[
        \sum_{t=0}^{\infty} \gamma^t \Big(
            \log r(s_t,a_t)
            + \tfrac{1}{(1-\gamma)\gamma^t} (
                \log \tfrac{p(s_{t+1} | s_t,a_t)}{q(s_{t+1} | s_t,a_t)}\\
                &\quad + \log \tfrac{\pi(a_t | s_t)}{\pi_c(a_t | s_t)}
            )
        \Big)
    \Big]\\
    &= (1-\gamma)\mathcal{L}(\pi)
\end{align*}

\end{proof}

\subsection{Classifiers Estimation Derivations} \label{Appendix: Classifiers Estimation Derivations}
We assume that all steps in the derivation are appropriate, that the denominator is non-zero, and that the argument of the logarithm is not zero.
\subsubsection{Transition Classifier Derivation.}
Define two events $\mathbb{E}$ and $\mathbb{M}$.

\begin{align*}
    \mathbb{E}=&\{\text{Given the state-action pair } (s_t,a_t), \text{ the agent interacts with} \\
    &  \text{the environmen }  p  \text{ to generate the subsequent state.} \}
\end{align*}
\begin{align*}
    \mathbb{M}=&\{\text{Given the state-action pair } (s_t,a_t), \text{ the agent interacts with} \\
    & \text{the model } m \text{ to generate the subsequent state.} \}
\end{align*}
The transition classifier is defined as $C_{\phi}(s_t,a_t,s_{t+1}) = P(\mathbb{E}|s_t,a_t,s_{t+1})$. Then
\begin{align}
    C_{\phi}(s,a,s_{t+1})=P(\mathbb{E}|s_t,a_t,s_{t+1})=&\frac{P(\mathbb{E},s_t,a_t)P(s_{t+1}|s_t,a_t,\mathbb{E})}{P(s_t,a_t,s_{t+1})} \nonumber\\
    &=\frac{P(\mathbb{E}|s_t,a_t)}{P(s_t,a_t)}\frac{P(s_{t+1}|s_t,a_t,\mathbb{E})}{P(s_t,a_t,s_{t+1})} .
    \label{transition classifier equation1}
\end{align} 
Events $\mathbb{E}$ and $\mathbb{M}$ form a partition of the probability space, which means that  $$P(\mathbb{M}|s_t,a_t,s_{t+1})=1-P(\mathbb{E}|s_t,a_t,s_{t+1})=1-C_{\phi}(s,a,s_{t+1}).$$
Similarly, we have
\begin{align}
    1-C_{\phi}(s,a,s_{t+1})&=\frac{P(\mathbb{M},s_t,a_t)P(s_{t+1}|s_t,a_t,\mathbb{M})}{P(s_t,a_t,s_{t+1})} \nonumber\\
    &=\frac{P(\mathbb{M}|s_t,a_t)}{P(s_t,a_t)}\frac{P(s_{t+1}|s_t,a_t,\mathbb{M})}{P(s_t,a_t,s_{t+1})}.
    \label{transition classifier equation2}
\end{align}
According to \eqref{transition classifier equation1} and \eqref{transition classifier equation2}, we have
\begin{align}
    \frac{C_{\phi}(s,a,s_{t+1})}{1-C_{\phi}(s,a,s_{t+1})} =\frac{P(s_{t+1}|s_t,a_t,\mathbb{E})}{P(s_{t+1}|s_t,a_t,\mathbb{M})}\frac{P(\mathbb{E}|s_t,a_t)}{P(\mathbb{M}|s_t,a_t)}.
    \label{transition classifier equation3}
\end{align}
Specifically, $P(s_{t+1}|s_t,a_t,\mathbb{E})=p(s_{t+1}|s_t,a_t)$, where $p$ represents the environment transition dynamic; $P(s_{t+1}|s_t,a_t,\mathbb{M})=m(s_{t+1}|s_t,a_t)$, where $m$ denotes the model transition dynamic. Then taking the logarithm of both sides of \eqref{transition classifier equation3}, we obtain:
\begin{equation}
    \log \frac{C_{\phi}(s,a,s_{t+1})}{1-C_{\phi}(s,a,s_{t+1})}=\log \frac{p(s_{t+1}|s_t,a_t)}{m(s_{t+1}|s_t,a_t)} 
 + \log \frac{P(\mathbb{E}|s_t,a_t)}{P(\mathbb{M}|s_t,a_t)}.
 \label{transition classifier equation4}
\end{equation}
Given the independence of the state-action pair $(s_t, a_t)$ from the transition dynamics, we assume that the visit distribution of $(s_t, a_t)$ remains identical. Then
\begin{equation}
    \frac{P(\mathbb{E}|s_t,a_t)}{P(\mathbb{M}|s_t,a_t)}  \approx \frac{d^{\pi}(s_t,a_t) \left| \mathcal{D}_{env} \right|}{d^{\pi}(s_t,a_t) \left| \mathcal{D}_{m} \right|}=\frac{\left| \mathcal{D}_{env} \right|}{\left| \mathcal{D}_{m} \right|},
    \label{transition classifier equation5}    
\end{equation}
where $\left| \mathcal{D}_{env} \right|$ and $\left| \mathcal{D}_{m} \right|$ represent the sizes of the datasets $\mathcal{D}_{env}$ and $\mathcal{D}_{m}$, respectively. Thus, $\frac{P(\mathbb{E}|s_t,a_t)}{P(\mathbb{M}|s_t,a_t)}$ can be approximated by a constant. Then according to \eqref{transition classifier equation4} and \eqref{transition classifier equation5}, we have
\begin{equation}
    \log \frac{p(s_{t+1}|s_t,a_t)}{m(s_{t+1}|s_t,a_t)} 
 \approx \log \frac{C_{\phi}(s,a,s_{t+1})}{1-C_{\phi}(s,a,s_{t+1})}-c_1,
 \nonumber
\end{equation}
where $c_1$ is a constant that does not influence the training of SAMBO.

\subsubsection{Action Classifier Derivation.}
The derivation of the action classifier is similar to that of the transition classifier.  
Define two events $\mathbb{A}$ and $\mathbb{C}$, where
$$
\mathbb{A} = \left\{\text{Given state } s_t, \text{ agent takes action } a_t \text{ according to policy } \pi \right\},
$$
$$
\mathbb{C} = \left\{\text{Given state } s_t, \text{ agent takes action } a_t \text{ according to policy } \pi_c \right\}.
$$
The action classifier is defined as $C_{\psi_1}(s_t,a_t) = P(\mathbb{A}|s_t,a_t)$, Then
\begin{align}
    C_{\psi_1}(s,a)=P(\mathbb{A}|s_t,a_t)=\frac{P(a_t|s_t,\mathbb{A})P(\mathbb{A}|s_t)P(s_t)}{P(s_t,a_t)}, \label{action classifier1} \\
    1- C_{\psi_1}(s,a)= P(\mathbb{C}|s_t,a_t)=\frac{P(a_t|s_t,\mathbb{C})P(\mathbb{C}|s_t)P(s_t)}{P(s_t,a_t)}.
    \label{action classifier2}
\end{align} 
Specifically, $P(a_t|s_t,\mathbb{A})=\pi(a_t | s_t)$, $P(a_t|s_t,\mathbb{C})=\pi_c(a_t | s_t)$. Then according to \eqref{action classifier1} and \eqref{action classifier2}, we have
\begin{align}
    \frac{C_{\psi_1}(s,a)}{1-C_{\psi_1}(s,a)} =\frac{\pi(a_t | s_t)}{\pi_c(a_t | s_t)}\frac{P(\mathbb{A}|s_t)}{P(\mathbb{C}|s_t)}.
    \label{action classifier3}
\end{align}
Similarly, $ \frac{P(\mathbb{A}|s_t)}{P(\mathbb{C}|s_t)}$ can also be approximated by a classifier $C_{\psi_2}(s)$, i.e. $ \frac{P(\mathbb{A}|s_t)}{P(\mathbb{C}|s_t)}=\frac{C_{\psi_2}(s)}{1-C_{\psi_2}(s)}$. Thus, we can approximate $\frac{\pi(a_t | s_t)}{\pi_c(a_t | s_t)}$ using two classifiers,
\begin{align}
\log \frac{\pi(a_t | s_t)}{\pi_c(a_t | s_t)} = \log \frac{C_{\psi_1}(s,a)}{1-C_{\psi_1}(s,a)} -  \log \frac{C_{\psi_2}(s)}{1-C_{\psi_2}(s)} .
\nonumber
\end{align}
For simplicity, we use the notation
\begin{align}
\log \frac{\pi(a_t | s_t)}{\pi_c(a_t | s_t)} = \log \frac{C_{\psi}(s,a)}{1-C_{\psi}(s,a)}.
\nonumber
\end{align}


\subsection{The Proof of Theorem \ref{TH: SAR is xi-uncertainty quantifier}} \label{Proof of Theorem 3}
\renewcommand{\thetheorem}{\ref{TH: SAR is xi-uncertainty quantifier}}
\begin{theorem} 
Under the assumption that $
D_{\operatorname{KL}}(m(\cdot \mid s, a) \,\|\, p(\cdot \mid s, a)) \geq \epsilon_{KL} >0
$, there exist $\alpha>0$ and $\beta>0$ such that the penalty of SAR forms a \(\xi\)-uncertainty quantifier.
\end{theorem}

\begin{proof}
	\begin{align*}
		&\quad \left|\widehat{\mathbb{B}}^{\pi}\widehat{V}_{t+1}(s,a)-\mathbb{B}^{\pi}\widehat{V}_{t+1}(s,a)\right| \\
		&=\gamma\left|\mathbb{E}_{s'\sim m}[\widehat{V}_{t+1}(s')]-\mathbb{E}_{s'\sim p}[\widehat{V}_{t+1}(s')]\right|\\
		&\leq \frac{\gamma r_{\max}}{1-\gamma}D_{\mathrm{TV}}(m(\cdot |s,a),p(\cdot |s,a))\\
		& \leq \frac{\gamma r_{\max}}{1-\gamma}  \sqrt{\frac{1}{2}D_{KL}(m(\cdot |s,a)||p(\cdot |s,a))} \\
		& \leq \frac{\gamma r_{\max}}{1-\gamma}  \frac{1}{\sqrt{2\epsilon_{KL}}} D_{KL}(m(\cdot |s,a)||p(\cdot |s,a)) \\
	\end{align*}
    
	The first inequality can be derived from the integral probability metric \cite{muller1997integral} associated with a class $\mathcal{F}={f: ||f||_{\infty} \leq 1}$.
	The second inequality can be derived from Pinsker's Inequality.
    The third inequality can be derived according to the assumption $
	D_{\operatorname{KL}}(m(\cdot \mid s, a) \,\|\, p(\cdot \mid s, a)) \geq \epsilon_{KL} >0$.
    
	For data in the model dataset $\mathcal{D}_{m}$, the dynamics are governed by the model $m$ and the data collection policy can be seen as $\pi$, \textit{i.e.} $\pi_c=\pi$. 
	Then the penalty of shifts-aware reward is $ D_{KL}(m(\cdot |s,a)||p(\cdot |s,a)) $. Thus, if $\alpha > 0$ is properly set, the penalty of SAR forms a \(\xi\)-uncertainty quantifier.
\end{proof}

\subsection{The Proof of Theorem \ref{TH: bound}}
Before proving Theorem \ref{TH: bound}, we introduce the pessimistic value iteration (PEVI) algorithm~\cite{jin2021pessimism}, a meta-algorithm for offline RL.

\begin{algorithm}[h]
\caption{Pessimistic Value Iteration (PEVI): General MDP}
\label{alg:pevi}
\begin{algorithmic}[1]
\Require Dataset $\mathcal{D} = \{(x_{\tau,h}^r, a_{\tau,h}^r, r_{\tau,h}^r)\}_{\tau,h}^{K,H}$.
\State Initialize $\hat{V}_{H+1}(\cdot) \leftarrow 0$.
\For{$h = H, H-1, \ldots, 1$}
    \State Construct $(\hat{\mathbb{B}}_h \hat{V}_{h+1})(\cdot, \cdot)$ and $\Gamma_h(\cdot, \cdot)$ based on $\mathcal{D}$.
    \State Set $\hat{Q}_h(\cdot, \cdot) \leftarrow (\hat{\mathbb{B}}_h \hat{V}_{h+1})(\cdot, \cdot) - \Gamma_h(\cdot, \cdot)$.
    \State Set $\hat{Q}_h(\cdot, \cdot) \leftarrow \min\{\hat{Q}_h(\cdot, \cdot), H - h + 1\}^+$.
    \State Set $\hat{\pi}_h(\cdot) \leftarrow \arg\max_{\pi_h(\cdot)} \langle \hat{Q}_h(\cdot, \pi_h(\cdot)), \pi_h(\cdot) \rangle_{\mathcal{A}}$.
    \State Set $\hat{V}_h(\cdot) \leftarrow \langle \hat{Q}_h(\cdot, \cdot), \hat{\pi}_h(\cdot) \rangle_{\mathcal{A}}$.
\EndFor
\State \textbf{Output}: $\mathrm{Pess}(\mathcal{D}) = \{\hat{\pi}_h\}_{h=1}^H$.
\end{algorithmic}
\end{algorithm}

The $\xi$-uncertainty quantifier characterizes the error between the empirical Bellman update and the true Bellman update. With the $\xi$-uncertainty quantifier, the following suboptimality theorem about Algorithm \ref{alg:pevi} holds.

\renewcommand{\thetheorem}{\ref{Def: xi-Uncertainty Quantifier}}
\begin{definition}
(\(\xi\)-Uncertainty Quantifier \cite{jin2021pessimism}).  
The set of penalization \(\{\Gamma_h\}_{h \in [H]}\) forms a \(\xi\)-uncertainty quantifier if it holds with probability at least \(1 - \xi\) that 
\[
\left| \widehat{\mathbb{B}} \widehat{V}_{t+1}(s, a) - \mathbb{B} \widehat{V}_{t+1}(s, a) \right| \leq \Gamma_t(s, a),   \quad  \forall(s,a)\in \mathcal{S} \times \mathcal{A}
\]
where \(\widehat{V}_{t+1}\) is an estimated value function at step \(t+1\) and \(\widehat{\mathbb{B}}\) is an empirical Bellman operator estimating the Bellman operator \(\mathbb{B}\). 
\end{definition}

\renewcommand{\thetheorem}{A}
\begin{theorem}[Suboptimality for PEVI \cite{jin2021pessimism}]
\label{Suboptimality for PEVI}
Suppose 
$\{ \Gamma_h \}_{h=1}^H$ in Algorithm \ref{alg:pevi} is a $\xi$-uncertainty quantifier. For any $s \in \mathcal{S}$, $\widehat{\pi}\in \text{Pess}(\mathcal{D})$ in Algorithm \ref{alg:pevi} satisfies
\[
\left| V^{\pi^*}(s_1) - V^{\widehat{\pi}}(s_1) \right|  \leq 2 \sum_{h=1}^H \mathbb{E}_{\pi^*} \big[ \Gamma_h(s_h, a_h) \mid s_1 = s \big].
\]

Here $\mathbb{E}_{\pi^*}$ is taken with respect to the trajectory induced by 
$\pi^*$ in the underlying MDP given the fixed function $\Gamma_h$.
\end{theorem}

\renewcommand{\thetheorem}{\ref{TH: bound}}
\begin{theorem}
Let $\pi^*$ be the optimal policy in the environment, $\pi^{*}_{G}$ be the optimal policy in MDP $G=(\mathcal{S},\mathcal{A},p,\log r,\mu_0,\gamma)$, $V_{G}^{\pi}(s)$ the value function in $G$ and $
\epsilon_{s} \geq D_{\operatorname{KL}}(m(\cdot \mid s, a) \,\|\, p(\cdot \mid s, a)) \geq \epsilon_{KL} >0
$.
Then the derived policy \(\widehat{\pi}\) according to SAR satisfies the following inequality
\begin{equation}
\left| V^{\pi^*}(s_1) - V^{\widehat{\pi}}(s_1) \right| 
\leq 2 \alpha H \epsilon_s+\epsilon(G;s_1),
\nonumber
\end{equation}
with probability at least \(1 - \xi\) for all \(s \in \mathcal{S}\). 
Here,
$$\epsilon(G;s)=|V^{\widehat{\pi}}(s) + V_{G}^{\pi^{*}_{G}}(s) -V^{\pi^*}(s) -V_{G}^{\widehat{\pi}}(s)  |,$$
which quantify the value error between different MDPs and $\epsilon(M;s)=0, \forall s \in \mathcal{S}$.
\end{theorem}

\begin{proof}
Decomposing the value bound,
\begin{equation}
     V^{\pi^*}(s) - V^{\widehat{\pi}}(s)=V^{\pi^*}(s) - V^{\widehat{\pi}}_G(s)+V^{\widehat{\pi}}_G(s)-V^{\pi^{*}_{G}}_G(s)+V^{\pi^{*}_{G}}_G(s)-V^{\widehat{\pi}}(s).
     \nonumber
\end{equation}
As established in the proof of Theorem 3, the penalty term employed by SAR serves as a $\xi$-uncertainty quantifier. Thus, according to Theorem \ref{Suboptimality for PEVI}, we have
\begin{equation}
    \left| V^{\widehat{\pi}}_G(s)-V^{\pi^{*}_{G}}_G(s) \right| 
\leq 2 \alpha \sum_{h=1}^{H} \mathbb{E}_{\pi^{*}_{G}} \left[ D_{KL}(m(\cdot |s_h,a_h)||p(\cdot |s_h,a_h)) \mid s_1 = s \right].
\nonumber
\end{equation}
Then 
\begin{align*}
    &\quad \left| V^{\pi^*}(s_1) - V^{\widehat{\pi}}(s_1) \right| \\
    &\leq 
    |V^{\widehat{\pi}}_G(s_1)-V^{\pi^{*}_{G}}_G(s_1)|+|V^{\pi^*}(s_1) - V^{\widehat{\pi}}_G(s_1)+V^{\pi^{*}_{G}}_G(s_1)-V^{\widehat{\pi}}(s_1)| \nonumber \\
    &\leq 2 \alpha \sum_{h=1}^{H} \mathbb{E}_{\pi^{*}_{G}} \left[ D_{KL}(m(\cdot |s_h,a_h)||p(\cdot |s_h,a_h)) \mid s_1 = s \right] + \epsilon(G;s_1) \nonumber\\
    & \leq 2 \alpha H \epsilon_s +\epsilon(G;s_1).
    \nonumber
\end{align*}
Specifically, if $\log r$ in $H$ is replaced by $r$, then $H=M$,$\pi^*_H=\pi^*$, $\epsilon(H;s)=\epsilon(M;s)=0, \forall s \in \mathcal{S}$.
\end{proof}

\section{Implementation Details} \label{Appendix: Implementation Details}
We present a detailed outline of SAMBO algorithm in Algorithm \ref{alg2}.

\subsection{Model Training}

In our approach, the model is represented by a probabilistic neural network that outputs a Gaussian distribution for both the next state and reward, given the current state and action:
$$
m_\theta\left(s_{t+1}, r_t \mid s_t, a_t\right)=\mathcal{N}\left(\mu_\theta\left(s_t, a_t\right), \Sigma_\theta\left(s_t, a_t\right)\right).
$$
Our model training approach is consistent with the methodology used in prior works~\cite{janner2019trust,yu2020mopo}.  
We train an ensemble of seven dynamics models and select the best five based on their validation prediction error from a held-out set containing 1000 transitions in the offline dataset $\mathcal{D}_{env}$. Each model in the ensemble is a 4-layer feedforward neural network with 200 hidden units. During model rollouts, we randomly choose one model from the best five models.

\begin{algorithm}[tb]
\caption{SAMBO-RL}
\label{alg2}
\textbf{Input}: Offline dataset $\mathcal{D}_{env}$, learned dyanmics models $\{m_{\theta}^i\}_{i=1}^N$, initialized policy $\pi$, shifts-aware reward hyperparameters $\alpha$ and $\beta$, rollout horizon $h$, rollout batch size $b$.
\begin{algorithmic}[1]
\State  Train the probabilistic dynamics model $m_{\theta}(s^{\prime},r | s,a)= \mathcal{N}(\mu_{\theta}(s,a) , \Sigma_{\theta}(s,a))$ on $\mathcal{D}_{env}$.
\State Initialize the model dataset buffer $\mathcal{D}_{m} \leftarrow \varnothing$. 
    \For{$i = 1,2,..., N_{iter}$}
        \State Initialize the policy dataset buffer $\mathcal{D}_{\pi} \leftarrow \varnothing$. 
        \For{$1,2,...,b$  (in parallel)}
            \State Sample state $s_1$ from $\mathcal{D}_{env}$ for the initialization of the rollout.
            \For{$j=1,2,...,h$}
                \State Sample an action $ a_j \sim \pi(s_j)$.
                \State Randomly pick dynamics $m$ from $\{m_{\theta}^i\}_{i=1}^N$ and sample $s_{j+1}, r_j \sim m(s_j,a_j)$.
                \State Add sample $(s_j,a_j,r_j,s_{j+1})$ to $\mathcal{D}_{m}$ and $\mathcal{D}_{\pi}$.
            \EndFor   
        \EndFor
        \State Update classifiers $C_{\phi}$ and $C_{\psi}$ according to Eq.~\eqref{classify_loss_pq} and Eq.~\eqref{classify_loss_pi}, respectively. 
        \State Drawing samples from $\mathcal{D}_{env} \cup \mathcal{D}_{m}$, compute shifts-aware reward according to Eq.~\eqref{pratical_reward_shift_model} and Eq.~\eqref{pratical_reward_shift_env}, respectively. 
        \State Run SAC with shifts-aware reward to update policy $\pi$.       
    \EndFor
\end{algorithmic}
\end{algorithm}

\subsection{Classifiers Training}
\label{Classifiers Training}

The transition classifier $C_{\phi}(s,a,s^{\prime}) \in [0,1]$ represents the probability that the transition originated from the environment. It is trained by minimizing the standard cross-entropy loss function:
\begin{align}
    \mathcal{L}_{\phi}=-\mathbb{E}_{\mathcal{D}_{env}} \left[ \log C_\phi(s,a,s^{\prime}) \right]-\mathbb{E}_{\mathcal{D}_{m}} \left[ \log (1-C_\phi(s,a,s^{\prime}))\right].
    \label{classify_loss_pq}
\end{align}
The action classifier $C_{\psi}(s,a) \in [0,1]$ represents the probability that a given state-action pair $(s,a)$ was generated by the current policy $\pi$. This classifier is trained by minimizing:
\begin{align}
    \begin{aligned}
        \mathcal{L}_{\psi_1} &= -\mathbb{E}_{\mathcal{D}_{\pi}} \left[ \log C_{\psi_1}(s,a) \right] - \mathbb{E}_{\mathcal{D}_{env}} \left[ \log (1-C_{\psi_1}(s,a))\right], \\
        \mathcal{L}_{\psi_2} &= -\mathbb{E}_{\mathcal{D}_{\pi}} \left[ \log C_{\psi_2}(s) \right] - \mathbb{E}_{\mathcal{D}_{env}} \left[ \log (1-C_{\psi_2}(s))\right].
    \end{aligned}
    \label{classify_loss_pi}
\end{align}
For data in the model dataset $\mathcal{D}_{m}$, the dynamics are governed by the model $m$ and the data collection policy can be seen as $\pi$, \textit{i.e.} $\pi_c=\pi$. 
Therefore, according to \eqref{reward_shift_practical in sambo}, $\forall (s,a,r,s^{\prime}) \in \mathcal{D}_{m}$, the shifts-aware reward simplifies to
\begin{align}
    \tilde{r}(s,a,s^{\prime})=\log r(s,a) + \alpha \log \frac{p(s^{\prime} | s,a)}{m(s^{\prime} | s,a)}.
    \nonumber
\end{align}
Then, for all $(s,a,s^{\prime}) \in \mathcal{D}_{m}$, the shift-aware reward can be estimated using the following formulation,
\begin{align}
    \tilde{r}(s,a,s^{\prime}) \approx \log r(s,a) + \alpha \log\frac{C_\phi(s,a,s^{\prime})}{1-C_\phi(s,a,s^{\prime})}.
    \label{pratical_reward_shift_model}
\end{align}

For data in the offline dataset $\mathcal{D}_{env}$, the dynamics are governed by environment $p$, and the training data collection policy aligns with the behavioral policy, \textit{i.e.} $\pi_c=\pi_b$. Therefore, $\forall (s,a,r,s^{\prime}) \in \mathcal{D}_{env}$, the shifts-aware reward simplifies to
\begin{align}
    \tilde{r}(s,a,s^{\prime})=\log r(s,a) + \beta \log\frac{\pi(a | s)}{\pi_b(a | s)},
    \nonumber
\end{align}
which is independent of the subsequent state $s^{\prime}$.
Then, for all $(s,a,s^{\prime}) \in \mathcal{D}_{env}$, the shift-aware reward can be estimated by
\begin{align}
    \tilde{r}(s,a,s^{\prime}) \approx \log r(s,a) + \beta \log\frac{C_\psi(s,a)}{1-C_\psi(s,a)}.
    \label{pratical_reward_shift_env}
\end{align}

\subsection{Policy Learning}
The policy optimization in our approach is based on the SAC algorithm, with hyperparameter configurations aligning with those employed in MOBILE~\cite{sun2023model}. For each update, we draw a batch of 256 transitions, with $5\%$ sourced from the offline dataset $ \mathcal{D}_{env} $ and the remaining $95\%$ from the synthetic dataset $ \mathcal{D}_{m} $.  The detailed hyperparameter settings used for the D4RL benchmark are provided in Table \ref{tab:hyperparameters of policy optimization}.

\begin{table}[ht]
    \centering
    \caption{Hyperparameters for policy optimization in SAMBO}
    \vskip 0.05in
    \begin{tabular}{lll}
        \hline
        \textbf{Hyperparameters} & \textbf{Value} & \textbf{Description} \\
        \hline
        K & 2 & The number of critics. \\
        $\gamma$ & 0.99 & Discount factor. \\
        $l_r$ of actor & $1 \times 10^{-4}$ & Policy learning rate. \\
        $l_r$ of critic & $3 \times 10^{-4}$ & Critic learning rate. \\
        Optimizer & Adam & Optimizers of the actor and critics. \\
        $f$ & 0.05 & Ratio of the real samples. \\
        Batch size & 256 & Batch size for each update. \\
        $N_{\text{iter}}$ & 3M & Total gradient steps. \\
        \hline
    \end{tabular}
    \label{tab:hyperparameters of policy optimization}
\end{table}

\section{Experimental Details} \label{Appendix: Experimental Details}

\subsection{Benchmark and Baselines}
We conduct experiments on Gym tasks (``v2'' version) within the D4RL~\cite{fu2020d4rl} benchmark. We also evaluated SAMBO on the NeoRL~\cite{qin2022neorl} benchmark, a more challenging dataset generated using a more conservative policy to better reflect real-world data collection scenarios. These datasets are characterized by their narrow scope and limited coverage.

\quad
        
\noindent
\textbf{D4RL.}We evaluate SAMBO against several baseline methods, including model-free methods such as BC~\cite{bain1995framework, ross2011reduction}, which learns the behavior policy, CQL~\cite{kumar2020conservative}, TD3+BC~\cite{fujimoto2021minimalist}, and EDAC~\cite{an2021uncertainty}, as well as model-based methods such as MOPO~\cite{yu2020mopo}, COMBO~\cite{yu2021combo}, TT~\cite{janner2021offline}, RAMBO~\cite{rigter2022rambo}, and MOBILE~\cite{sun2023model}. We now provide the sources for the reported performance on this benchmark. For TD3+BC~\cite{fujimoto2021minimalist}, EDAC~\cite{an2021uncertainty}, TT~\cite{janner2021offline}, RAMBO~\cite{rigter2022rambo}, and MOBILE~\cite{sun2023model}, we cite the scores reported in the original papers, as these evaluations were conducted on the ``v2" datasets in Gym. 
Since COMBO~\cite{yu2021combo} does not provide source code, we include the results as reported in the original paper. The results of CQL~\cite{kumar2020conservative} are taken from the performance table in MOBILE, where CQL was retrained on the ``v2" datasets in Gym. The results of MOPO* are obtained from experiments using the OfflineRL-Kit library\footnote{https://github.com/yihaosun1124/OfflineRL-Kit} on the ``v2" datasets. These scores are referenced in Table \ref{table: score in d4rl}.


\quad
        
\noindent
\textbf{NeoRL.}We compare SAMBO with baseline methods that align with the d4rl experiments, excluding TT, COMBO, and RAMBO, as no results for these methods are reported in their original papers or the NeoRL paper, and determining suitable hyperparameters for them would require an excessive amount of computational time. We report the performance of BC, CQL, and MOPO based on the original NeoRL paper, while the performance of TD3+BC, EDAC, and MOBILE is derived from the scores in the MOBILE paper.


\subsection{Hyperparameters} \label{Appendix:Hyperparameters}
We list the hyperparameters that have been tuned as follows. The specific settings are detailed in Table \ref{tab:hyperparameters in SAMBO in d4rl} and Table \ref{tab:hyperparameters in SAMBO in neorl}.

\begin{table}[htbp]
    \centering
    \caption{Hyperparameters of SAMBO used in the D4RL.}
    \vskip 0.05in
    \begin{tabular}{lccc}
        \toprule
        \textbf{Task Name} & $\boldsymbol{\alpha}$ & $\boldsymbol{\beta}$ & $\boldsymbol{h}$  \\
        \midrule
        \textbf{halfcheetah-random}      & 0.01 & 0.01 & 5 \\
        \textbf{hopper-random}           & 0.02 & 0.01 & 5 \\
        \textbf{walker2d-random}         & 0.3 & 0.3 & 5 \\
        \midrule
        \textbf{halfcheetah-medium}      & 0.03 & 0.03 & 5 \\
        \textbf{hopper-medium}           & 0.02 & 0.01 & 5 \\
        \textbf{walker2d-medium}        & 0.03 & 0.01 & 5 \\
        \midrule
        \textbf{halfcheetah-medium-replay} & 0.01 & 0.01 & 5 \\
        \textbf{hopper-medium-replay}    & 0.05 & 0.05 & 5 \\
        \textbf{walker2d-medium-replay}  & 1 & 1 & 1 \\
        \midrule
        \textbf{halfcheetah-medium-expert} & 0.3 & 0.3 & 5 \\
        \textbf{hopper-medium-expert}    & 1 & 1 & 5 \\
        \textbf{walker2d-medium-expert}  & 2 & 2 & 1 \\
        \bottomrule    
    \end{tabular}
    \label{tab:hyperparameters in SAMBO in d4rl}
\end{table}

\begin{table}[ht]
    \centering
    \caption{Hyperparameters of SAMBO used in the NeoRL.}
    \vskip 0.05in
    \begin{tabular}{lcccc}
        \toprule
        \textbf{Task Name} & $\boldsymbol{\alpha}$ & $\boldsymbol{\beta}$ & $\boldsymbol{h}$ \\
        \midrule
        \textbf{Halfcheetah-L}      & 0.1 & 0.1 & 5 \\
        \textbf{Hopper-L}           & 0.3 & 0.3 & 5 \\
        \textbf{Walker2d-L}         & 3 & 1 & 1 \\
        \midrule
        \textbf{Halfcheetah-M}      & 0.1 & 0.1 & 5 \\
        \textbf{Hopper-M}           & 0.3 & 0.3 & 5 \\
        \textbf{Walker2d-M}        & 3 & 1 & 1 \\
        \midrule
        \textbf{Halfcheetah-H} & 5 & 1 & 5 \\
        \textbf{Hopper-H}    & 0.3 & 0.3 & 5 \\
        \textbf{Walker2d-H}  & 3 & 1 & 1 \\
        \bottomrule    
    \end{tabular}
    \label{tab:hyperparameters in SAMBO in neorl}
\end{table}
      
\noindent
\textbf{Coefficient $\alpha$ and $\beta$.} 
The model bias coefficient, $\alpha$, determines the degree of correction for model bias, while the policy shift coefficient, $\beta$, regulates the extent of correction for policy shift.

\quad
        
\noindent
\textbf{Rollout Length $h$.} Similar to MOPO, We perform short-horizon branch rollouts in SAMBO. we tune $h$ in the
range of $\{ 1,5 \}$ for Gym tasks in the D4RL benchmark.
We terminate training at 1M to prevent potential performance degradation due to the rapid convergence of the HalfCheetah environment in NeoRL. Additionally, to maintain training stability in the HalfCheetah-high dataset, we freeze classifiers after 0.5 million iterations. Similarly, in the walker2d-medium dataset of the D4RL benchmark, we ensure training stability by fixing the transition classifier after 1 million iterations.

\subsection{Reward Function Discussion}

Based on our derivation, if we start with $\mathbb{E}_{p^{\pi}(\tau)} e^{R(\tau)}$, we can obtain a new surrogate objective, which we refer to as SAMBO-r. In scenarios with sparse rewards and narrow datasets, where $\log r$ may distort the reward, SAMBO-r may be more suitable as a surrogate objective.
\begin{itemize}
    \item \textbf{SAMBO.} The modified reward is
    $$\tilde{r}(s,a,s^{\prime}) = \log r^{\prime}(s,a) + \alpha \log\frac{C_\phi(s,a,s^{\prime})}{1-C_\phi(s,a,s^{\prime})} + \beta \log\frac{C_\psi(s,a)}{1-C_\psi(s,a)}.$$
    \item \textbf{SAMBO-r.} The modified reward is
    $$\tilde{r}(s,a,s^{\prime}) =  r^{\prime}(s,a) + \alpha \log\frac{C_\phi(s,a,s^{\prime})}{1-C_\phi(s,a,s^{\prime})} + \beta \log\frac{C_\psi(s,a)}{1-C_\psi(s,a)}.$$
    In this case, $\epsilon(G;s)=0, \forall s \in \mathcal{S}$ in Theorem \ref{TH: bound}, but this derived surrogate objective is not a lower bound of the original RL objective.
\end{itemize}

\subsection{Computing Infrastructure} \label{Appendix:Computing Infrastructure}
All experiments are conducted using a single GeForce GTX 3090 GPU and an Intel(R) Xeon(R) Gold 6330 CPU @ 2.00GHz. Our implementation code is based on the OfflineRL-Kit library. 
The software libraries and frameworks utilized in SAMBO are consistent with those specified in this library.

\section{Omitted Experiments}

\subsection{Visualization}

\begin{figure}[ht]
\vskip -0.1in
    \centering
    \includegraphics[width=0.4\columnwidth]{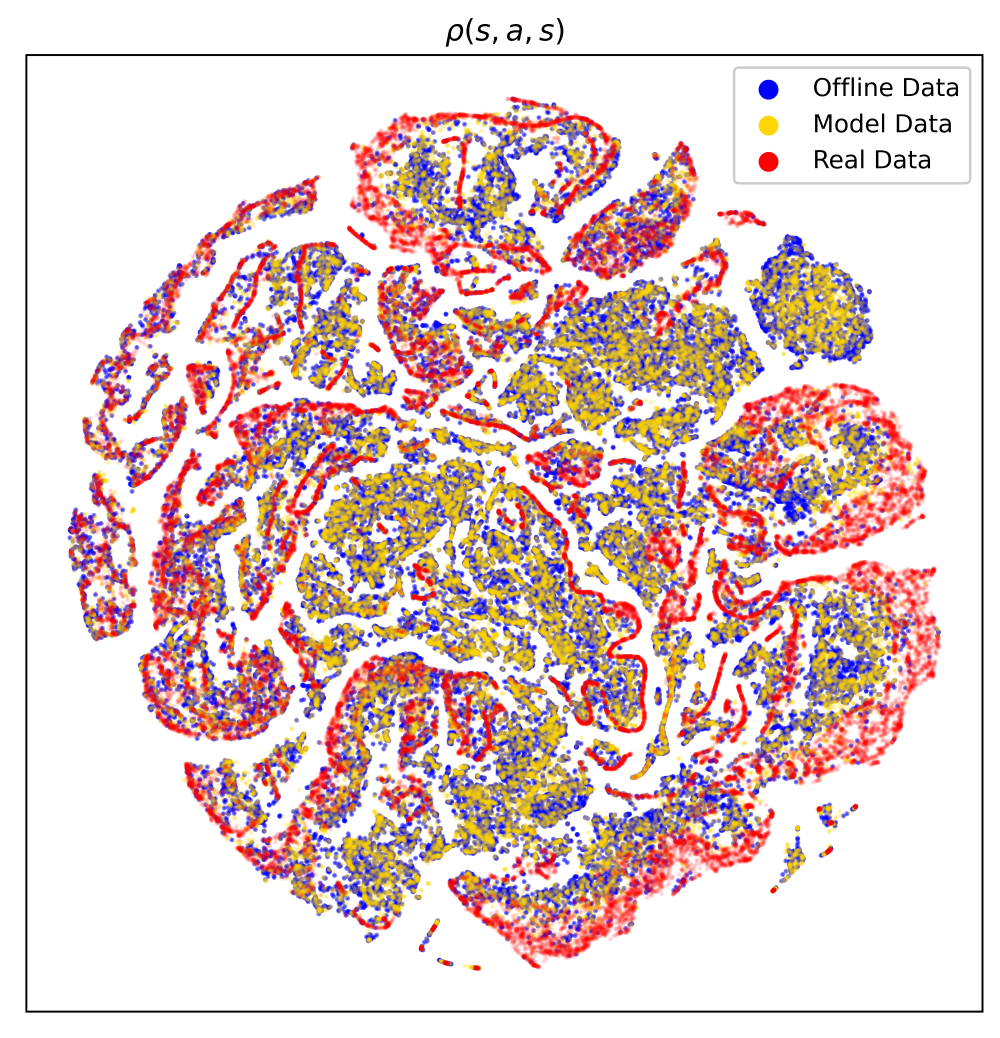}
    \caption{The $(s, a, s^{\prime})$ distributions generated by SAMBO in the walker2d-medium-expert task in D4RL benchmark. Model Data represents the distribution of data generated by executing the final learned policy in the dynamic model, while Real Data represents the distribution of data generated by executing the final learned policy in the environment.}
    \label{fig: distribution visualization in walker-me}  
\end{figure}

We also conducted a visualization experiment on the walker2d-medium-expert dataset within the D4RL benchmark. As shown in Figure \ref{fig: distribution visualization in walker-me}, SAMBO effectively constrains the Model Data to closely match the Offline Data, and the Real Data generated by executing the learned policy in the actual environment does not significantly deviate from either the Offline Data or Model Data. This highlights the effectiveness of SAMBO in mitigating distribution shift across both benchmarks.

\begin{figure}[t]
    \vskip -0.1in
    \centering
    \begin{minipage}[t]{0.45\linewidth}
        \centering
        \includegraphics[width=\linewidth]{figure/Appendix/H-l.png}
        \caption{ Hopper-L}
        \label{fig: Performance compare in Hopper-L}
    \end{minipage}%
    \hfill
    \begin{minipage}[t]{0.45\linewidth}
        \centering
        \includegraphics[width=\linewidth]{figure/Appendix/H-m.png}
        \caption{ Hopper-M}
    \end{minipage}

    \begin{minipage}[t]{0.45\linewidth}
        \centering
        \includegraphics[width=\linewidth]{figure/Appendix/H-h.png}
        \caption{ Hopper-H}
    \end{minipage}%
    \hfill
    \begin{minipage}[t]{0.45\linewidth}
        \centering
        \includegraphics[width=\linewidth]{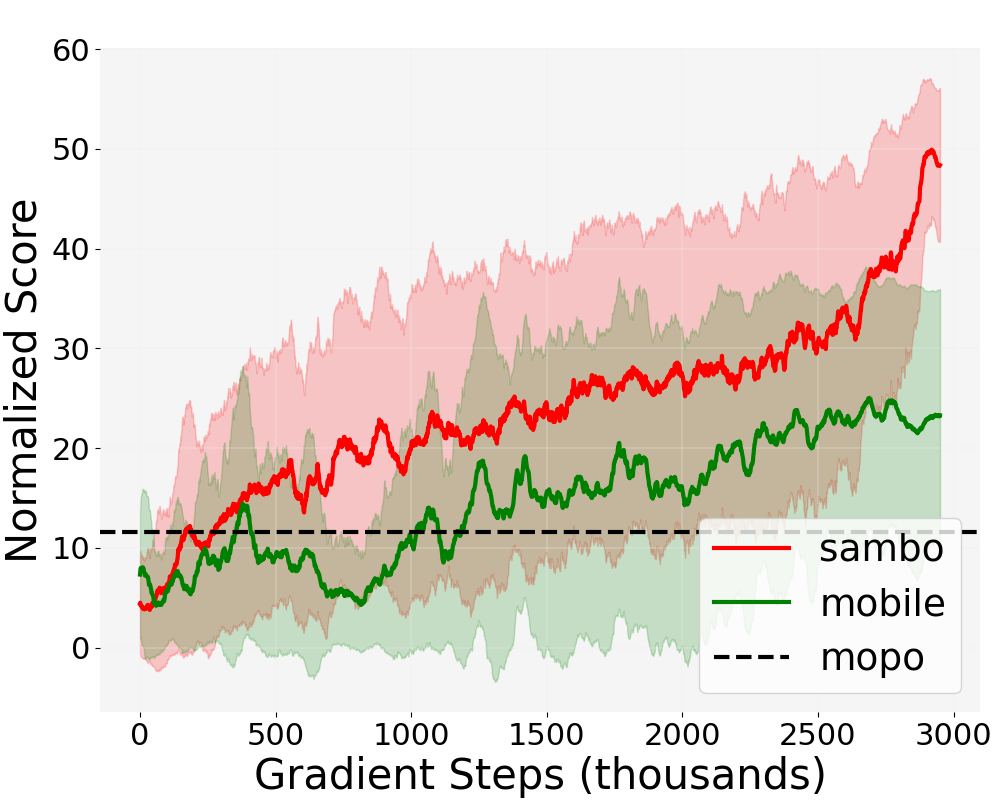}
        \caption{ Walker2d-L}
    \end{minipage}

    \begin{minipage}[t]{0.45\linewidth}
        \centering
        \includegraphics[width=\linewidth]{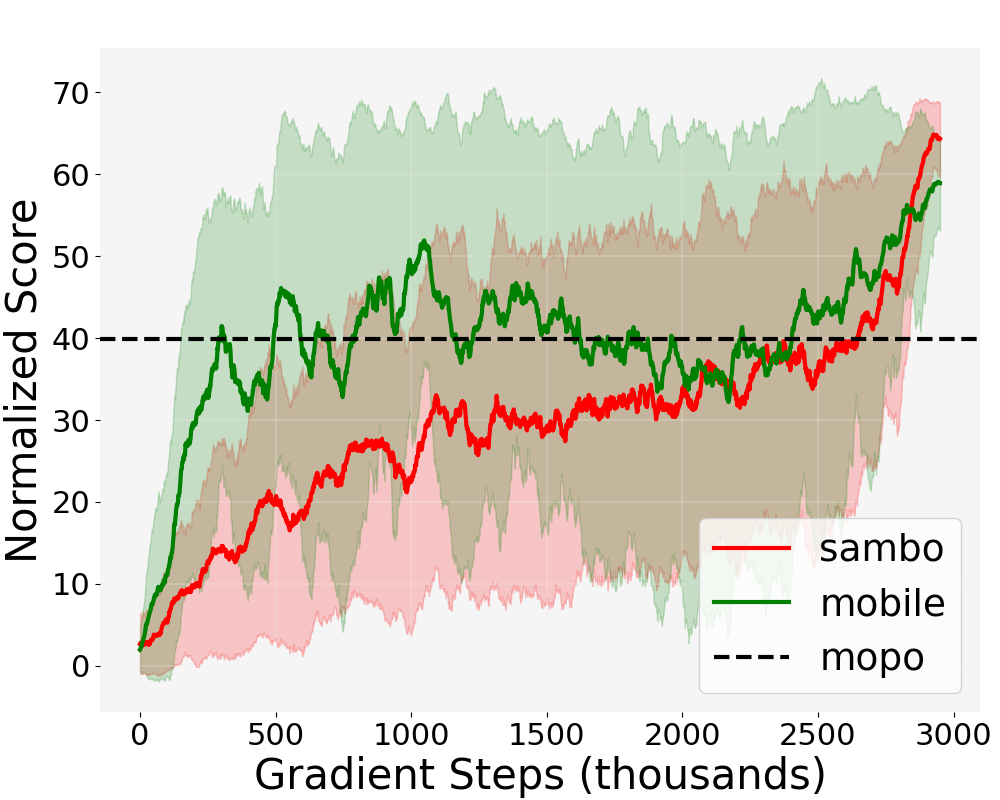}
        \caption{ Walker2d-M}
    \end{minipage}%
    \hfill
    \begin{minipage}[t]{0.45\linewidth}
        \centering
        \includegraphics[width=\linewidth]{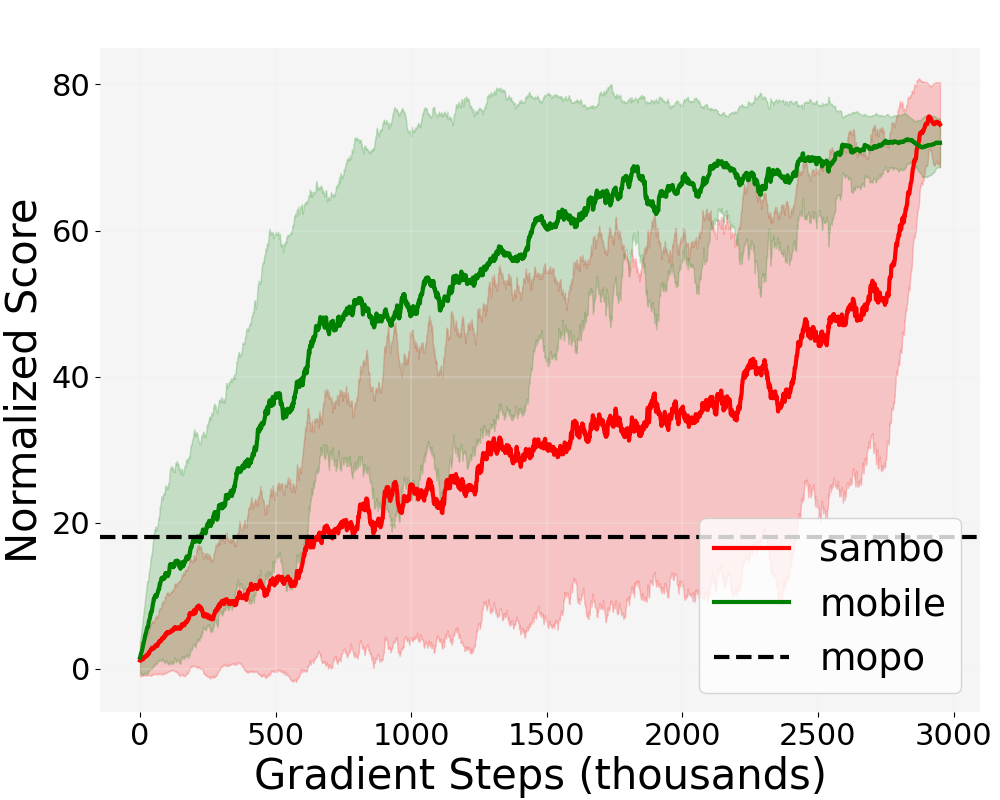}
        \caption{ Walker2d-H}
    \end{minipage}

    \begin{minipage}[t]{0.45\linewidth}
        \centering
        \includegraphics[width=\linewidth]{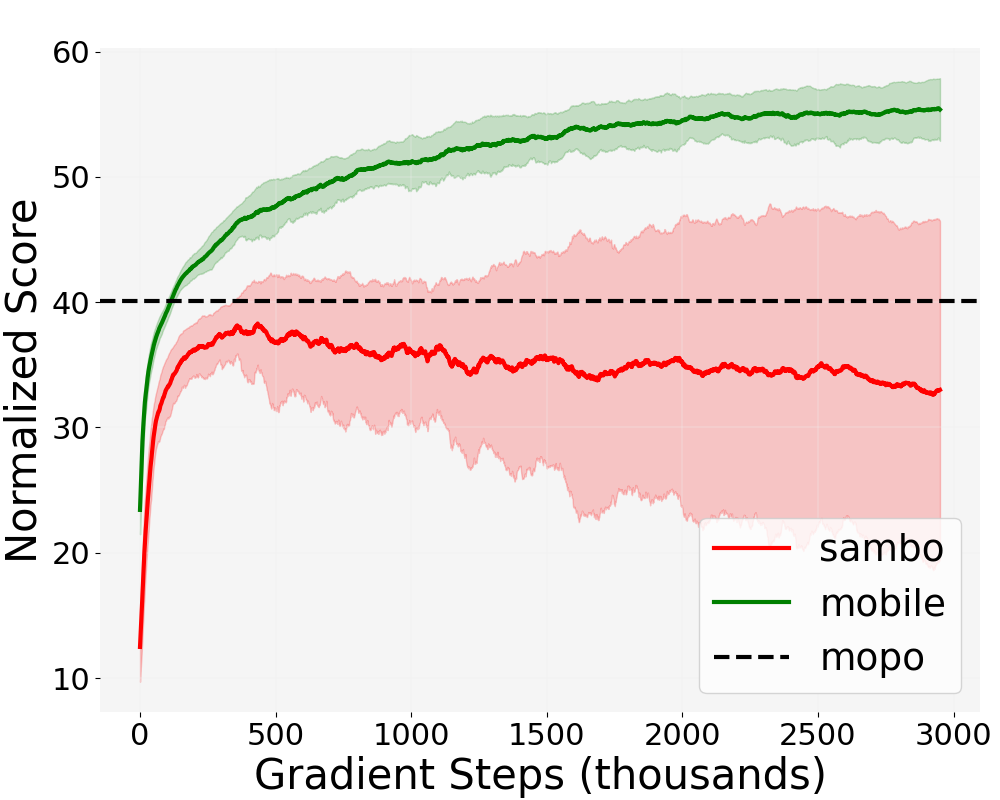}
        \caption{ HalfCheetah-L}
    \end{minipage}%
    \hfill
    \begin{minipage}[t]{0.45\linewidth}
        \centering
        \includegraphics[width=\linewidth]{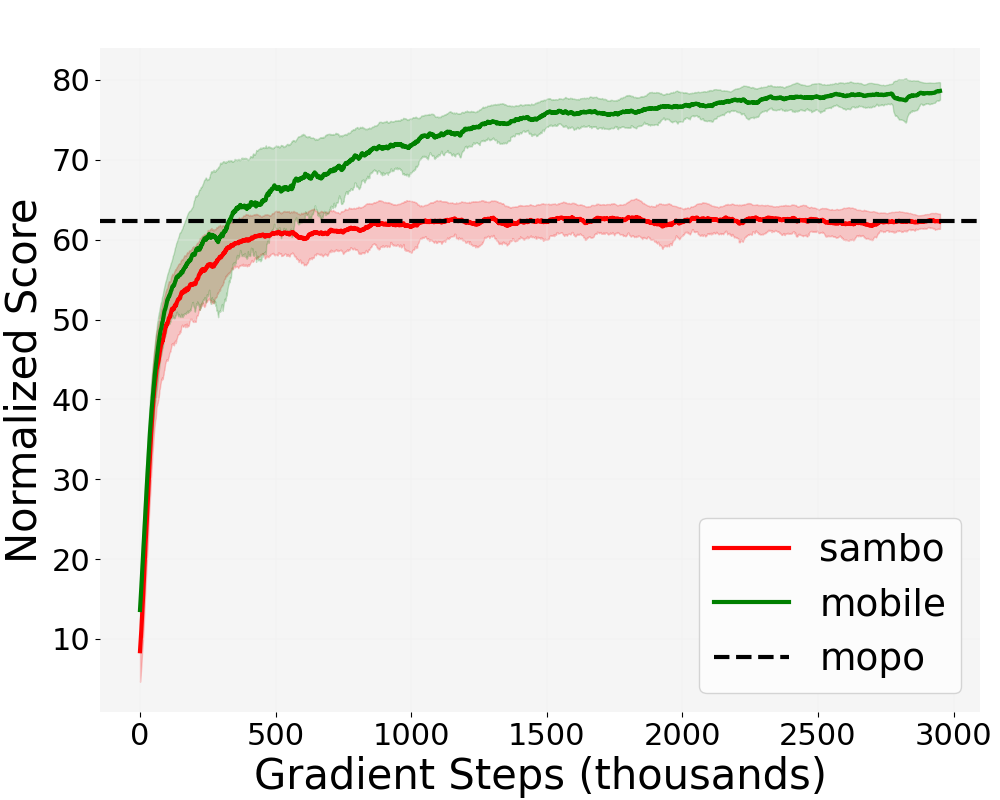}
        \caption{ HalfCheetah-M}
    \end{minipage}

    \begin{minipage}[t]{0.45\linewidth}
        \centering
        \includegraphics[width=\linewidth]{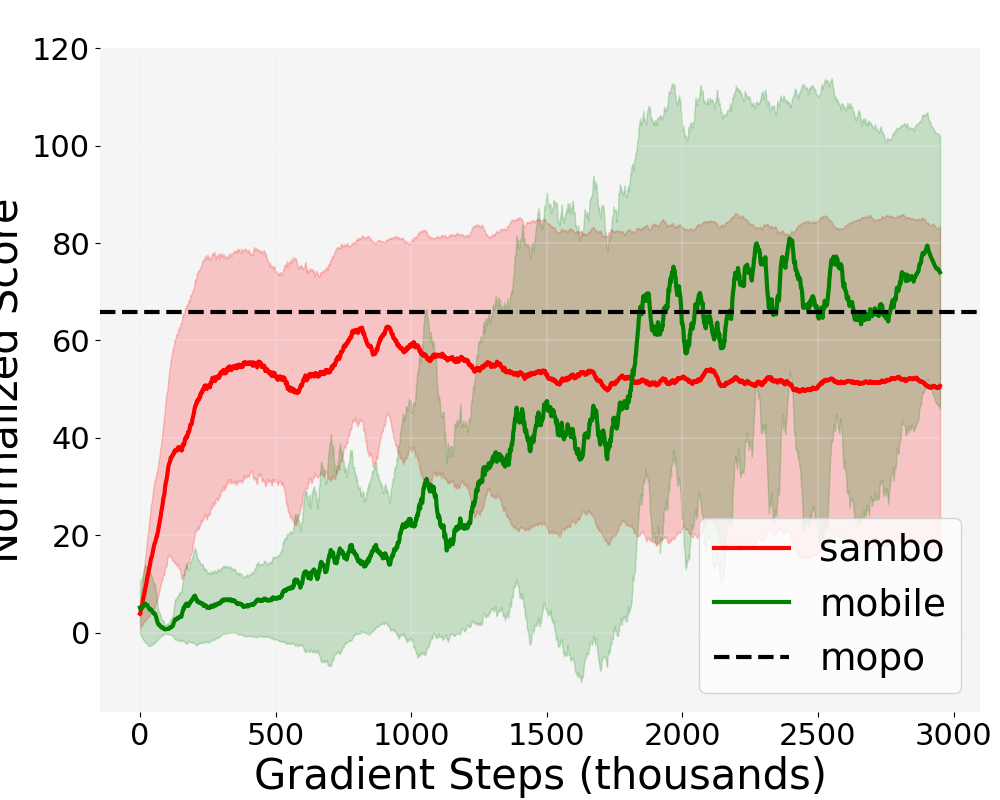}
        \caption{ HalfCheetah-H}
        \label{fig: Performance compare in HalfCheetah-H}
    \end{minipage}
\end{figure}

\subsection{Performance Comparison} \label{Appendix: Performance Compare}

We also compared the performance curves of model-based approaches on other tasks in the NeoRL benchmark. The curve for MOBILE was obtained by executing the official implementation. Since MOPO does not have an official implementation on NeoRL, we report its score based on the results presented in the original NeoRL paper.
As shown in Figures \ref{fig: Performance compare in Hopper-L} to \ref{fig: Performance compare in HalfCheetah-H}, SAMBO outperforms in all Hopper and Walker2d tasks, both of which are more complex and challenging. Although it exhibits slower convergence initially in the Walker2d-M and Walker2d-H tasks, its performance improves rapidly as the experiment progresses, ultimately achieving SOTA levels. 
On the other hand, SAMBO does not achieve strong performance on the HalfCheetah tasks. We attribute this to the fact that the HalfCheetah score mainly depends on achieving high running speed rather than mere stability. Since SAMBO’s dual conservatism with respect to model bias and policy shift places greater emphasis on stability, it performs less competitively on HalfCheetah, while demonstrating strong results on Hopper and Walker2d, where stability is more critical.

\end{document}